\documentclass[preprint,12pt]{elsarticle}

\usepackage{amssymb}
\usepackage{amsmath}

\usepackage{color}
\usepackage{pifont}

\usepackage{wrapfig}
\usepackage{multirow}
\usepackage{stfloats}
\usepackage{graphicx, subfig}
\usepackage[normalem]{ulem}
\useunder{\uline}{\ul}{}
\usepackage{enumitem}

\journal{Neural Networks}

\begin{document}

\begin{frontmatter}



\title{Bridging Graph Structure and Knowledge-Guided Editing for Interpretable Temporal Knowledge Graph Reasoning}

\author[label1]{Shiqi Fan}
\author[label3]{Quanming Yao}
\author[label4]{Hongyi Nie}
\author[label5]{Wentao Ma}
\author[label2]{Zhen Wang}
\author[label1]{Wen Hua\corref{cor1}}
\ead{wency.hua@polyu.edu.hk}
\affiliation[label1]{organization={Department of Data Science and Artificial Intelligence, The Hong Kong Polytechnic University},
            city={Hong Kong},
            postcode={999077},
            country={China}}

\affiliation[label2]{organization={School of Cybersecurity, Northwestern Polytechnical University},
            city={Xi’an},
            postcode={710072},
            state={Shaanxi},
            country={China}}

\affiliation[label3]{organization={Department of Electronic Engineering, Tsinghua University},
            city={BeiJing},
            postcode={100084},
            country={China}}

\affiliation[label4]{organization={School of Mechanical Engineering, Northwestern Polytechnical University},
            city={Xi’an},
            postcode={710072},
            state={Shaanxi},
            country={China}}

\affiliation[label5]{organization={School of Mathematics and Statistics, Xi’an Jiaotong University},
            city={Xi’an},
            postcode={710049},
            state={Shaanxi},
            country={China}}

\cortext[cor1]{Corresponding author.}



\begin{abstract}
Temporal knowledge graph reasoning (TKGR) aims to predict future events by inferring missing entities with dynamic knowledge structures. Existing LLM-based  reasoning methods prioritize contextual over structural relations, struggling to extract relevant subgraphs from dynamic graphs. This limits structural information understanding, 
leading to unstructured, hallucination-prone inferences especially with temporal inconsistencies. To address this problem, we propose IGETR (Integration of Graph and Editing-enhanced Temporal Reasoning), a hybrid reasoning framework that combines the structured temporal modeling capabilities of Graph Neural Networks (GNNs) with the contextual understanding of LLMs. IGETR operates through a three-stage pipeline. The first stage aims to ground the reasoning process in the actual data by identifying structurally and temporally coherent candidate paths through a temporal GNN, ensuring that inference starts from reliable graph-based evidence. The second stage introduces LLM-guided path editing to address logical and semantic inconsistencies, leveraging external knowledge to refine and enhance the initial paths. The final stage focuses on integrating the refined reasoning paths to produce predictions that are both accurate and interpretable. {Experiments on standard TKG benchmarks show that IGETR achieves state-of-the-art performance, outperforming strong baselines with relative improvements of up to 5.6\% on Hits@1 and 8.1\% on Hits@3 on the challenging ICEWS datasets.} Additionally, we execute ablation studies and additional analyses confirm the effectiveness of each component.
\end{abstract}




\end{frontmatter}



\section{Introduction}
\label{sec1}
Temporal Knowledge Graphs (TKGs) extend traditional knowledge graphs by explicitly integrating temporal information, thereby representing dynamic relationships and evolving patterns among entities more accurately~\cite{10577554, ji2021survey, liang2150better, guo2025semantic}. 
This temporal modeling enables a detailed analysis of relational dynamics and supports inference on future events, effectively capturing the underlying evolutionary and chronological dependencies~\cite{li2021temporal, zhang2023few}. {The challenge of reasoning over such dynamic graph structures is not unique. Graph Neural Networks (GNNs) have emerged as a powerful paradigm for modeling complex, evolving relationships in diverse domains, ranging from urban traffic flow prediction~\cite{B,D} to human action recognition~\cite{E}. Their success in these applications motivates our choice to employ GNNs in this work.} 
Consequently, TKGs have demonstrated significant applicability across diverse fields, including recommendation systems~\cite{qin2023learning, huang2024knowledge}, real-time question answering~\cite{liu2023prompt, yang2025language}, and event reasoning~\cite{cai2024survey, li2025temporal}. {The successful application of reasoning in these areas mirrors its importance in other complex networked systems, such as the Internet of Things, where intelligent models are critical for challenges like resource management~\cite{RefA} and attack detection~\cite{RefB}.}

The fundamental challenge in TKG reasoning (TKGR) lies in achieving trustworthy and accurate reasoning, where both the predicted results and the inference paths must be logical, interpretable, and temporally coherent. As real-world events unfold dynamically, it is crucial not only to ensure prediction accuracy but also to generate reasoning chains that adhere to chronological constraints and causal structures, for instance, trade agreements logically follow negotiations rather than precede them. However, existing methods struggle to balance predictive performance with explainability~\cite{luo2023chatrule}. {The widely adopted graph-based methods (e.g., xERTE~\cite{xerte}, CaORG~\cite{fan2024flow}, RPC~\cite{liang2023learn}), excel at capturing structural dependencies from observed data but are fundamentally data-bound. The reasoning they support is constrained by the existing graph structure, reflecting statistical patterns rather than deep logical coherence~\cite{dai2022mrgat, ma2024harnessing}. Critically, they lack a mechanism for external knowledge integration, making them unable to correct erroneous links or infer plausible paths beyond the observed data~\cite{zhang2024can, guo2024graphedit}.}
 Conversely, Large Language Models (LLMs), with their vast background knowledge and strong reasoning capabilities, provide an alternative approach to TKGR (e.g., GenTKG~\cite{liao2024gentkg}, ICL~\cite{ICL2023temporal}, COH~\cite{luo2024chain}). However, their application to temporal reasoning is hindered by several critical issues. First, they face challenges adapting to dynamic environments, as it is impractical to frequently update LLMs to incorporate evolving knowledge from TKGs. Even with approaches like in-context learning, extracting relevant subgraphs and effectively understanding structured information from dynamic graphs remain challenging.~\cite{han2024parameter, maltoni2024arithmetic}. Second, LLMs suffer from hallucination, often generating plausible but chronologically inconsistent inference paths due to outdated or conflicting knowledge~\cite{pan2024unifying}. Third, they lack controllable explainability, as their reasoning outputs are not constrained by structured graph evidence, making paths difficult to be interpreted and reproduced~\cite{dwivedi2023explainable}. {While hybrid approaches such as Retrieval-Augmented Generation (RAG) \cite{gao2023retrieval, feng2025retrieval} and Re-ranking \cite{wang2024kc, chu2024timebench} have been explored, they fundamentally differ from our objective. These paradigms typically treat the KG as a static repository to augment LLM generation, or passively discriminate among retrieved candidates. Consequently, they risk propagating retrieval noise or logical gaps into the final inference. A critical gap remains: the lack of a mechanism that can actively validate and rectify the structural reasoning paths themselves.}

To bridge this gap, how to integrate the model from both the data and knowledge perspectives to complement each other's strengths constitutes the core motivation of our work. We propose \textbf{I}ntegration of \textbf{G}raph and \textbf{E}diting-enhanced \textbf{T}emporal \textbf{R}easoning (IGETR), a hybrid framework that strategically combines \textit{the structural reasoning capacity} of GNNs with \textit{the contextual understanding} of LLMs, ensuring that reasoning remains both \textit{data-driven} and \textit{knowledge-guided}. Our approach constructs a three-stage pipeline that progressively refines reasoning paths to improve prediction accuracy, interpretability, and temporal coherence. Initially, we employ a temporal GNN with attention edge sampling to extract reasoning chains directly from the TKG, prioritizing chronologically proximal connections while maintaining diversity through stratified attention mechanisms. This ensures that the generated paths are structurally and temporally aligned with the underlying data. Next, we introduce an LLM-mediated path editing mechanism, which reviews and refines the extracted inference chains, leveraging background knowledge to correct inconsistencies, resolve ambiguities, and enhance logical coherence. By doing so, we address the limitations of purely data-driven methods, allowing for more flexible and knowledge-informed reasoning. Finally, we integrate a graph Transformer module that dynamically weights the edited paths based on their temporal relevance and structural coherence, enabling adaptive fusion of multi-hop evidence while preserving explainability. Unlike traditional graph-based methods that are restricted to learned statistical patterns or LLM-based approaches that generate unconstrained inferences, our framework maintains the interpretability of graph models while incorporating the contextual depth of LLMs, resulting in a more robust and trustworthy reasoning paradigm for TKGR.

By combining the structured learning capabilities of GNNs with the reasoning flexibility of LLMs, IGETR addresses key challenges in temporal reasoning, ensuring that predictions remain not only accurate but also logically interpretable and adaptable to real-world dynamics. This novel integration enhances the reliability of reasoning paths, making it particularly suitable for high-stakes applications where both prediction performance and transparent justifications are paramount. Our contributions can be summarized as follows:

\begin{itemize}[leftmargin=*]
\item[$\bullet$]
We propose IGETR, the first path-refine reasoning framework integrating temporal graph neural networks with knowledge-augmented LLMs, effectively combining structural reasoning and semantic refinement to improve logical consistency and interpretability.

\item[$\bullet$]
We design a three-stage pipeline that grounds reasoning in graph structures,  refines paths via LLM editing, and aggregates multi-hop evidence with a graph Transformer, ensuring data-driven reliability while enabling knowledge-guided, controllable reasoning.

\item[$\bullet$]
We validate the superiority of IGETR through experiments on three TKG datasets, demonstrating its effectiveness in addressing the key challenges of TKGR, with ablation studies validate the effectiveness of key components in the framework.

\end{itemize}

\section{Related Work}

\subsection{Temporal Knowledge Graph Reasoning}

Early methods for Temporal Knowledge Graph Reasoning (TKGR) predominantly adopt embedding-based strategies, where entities and relations are projected into continuous vector spaces. These models treat temporal knowledge as transformations in time-dependent embedding spaces. For instance, TTransE~\cite{ttranse}, TNTComplEx~\cite{TNTCOMP}, and ChronoR~\cite{ChronoR} extend static models by incorporating temporal representations into their scoring functions. TeLM~\cite{xu-etal-2021-telm} introduces temporal logic into embedding composition, while TANGO~\cite{han-etal-2021-tango} models continuous-time evolution using neural ODEs. Although these methods are efficient and scalable, they often fail to capture complex structural dependencies or provide interpretable reasoning, making them less effective in forecasting tasks that require temporal and causal reasoning.

Graph-based approaches have since become a dominant paradigm for TKGR due to their ability to model structural and temporal dependencies explicitly. RE-NET~\cite{jin2020recurrent} combines recurrent event encoders with neighborhood aggregation to capture dynamic evolution, while RE-GCN~\cite{li2021temporal} leverages graph convolutions and GRUs to model sequential influence across time. xERTE~\cite{xerte} proposes attention-based subgraph extraction to isolate relevant local structures for each query, and TANGO~\cite{han2021learning} introduces temporal neighborhood sampling for more robust inference. TiRGN~\cite{li2022tirgn} and HGLS~\cite{zhang2023learning} further improve long-range dependency modeling through temporal trajectory learning. CaORG~\cite{fan2024flow} introduces candidate-oriented reasoning to enhance relevance and reduce noise in graph traversal. {Also, this trend mirrors successes in broader domains, where dynamic spatio-temporal GNNs effectively model systems like citywide traffic flows~\cite{A, F} and perform complex resource allocation~\cite{G}}.

Despite their effectiveness, these models primarily rely on statistical patterns in observed data and are constrained by the fixed structure of the graph. This limitation reduces their ability to hypothesize new relations, rectify noisy edges, or generalize beyond the observed subgraph. Consequently, while graph-based methods capture global structure well, they may struggle in complex, evolving environments where reasoning requires flexibility, external context, and logical consistency across time.

\subsection{Large Language Models for Graph}
Recent advancements in LLMs have significantly reshaped KG reasoning by leveraging extensive pre-trained knowledge and robust reasoning capabilities. Integration strategies between KGs and LLMs have proven effective in enhancing tasks such as link prediction, knowledge completion, and logical inference. For instance, StructGPT~\cite{jiang2023structgpt} employs iterative reasoning refinement using structural KG information, effectively improving complex reasoning tasks. Instruction-tuning methods~\cite{zhang2023instruction} enhance KG reasoning capabilities through prompt engineering and fine-tuning strategies, with notable scalability benefits. Retrieval-based approaches, including KICGPT~\cite{wei2023kicgpt} and KC-GenRe~\cite{wang2024kc}, demonstrate substantial accuracy improvements by combining KG retrievers with LLM-based re-ranking mechanisms. Methods such as MPiKGC~\cite{xu2024multi} utilize internal LLM knowledge to refine smaller KG embedding models, illustrating the complementary strengths of structured graph representations and textual embeddings. Additionally, KG-LLM~\cite{shu2024knowledge} and PROLINK~\cite{wang2024llm} translate structured paths into textual contexts or prompt graph networks, significantly enhancing the interpretability and controllability of reasoning processes. However, despite these successes, challenges remain, including logical hallucinations, uncontrollable reasoning paths, and difficulties adapting rapidly to evolving knowledge.

Recently, several methods such as Graph4GPT~\cite{guo2023gpt4graph} and GraphLLM~\cite{chai2023graphllm} specifically encode graph structures into sequential inputs for LLMs, further demonstrating their potential for structured reasoning. Nevertheless, these methods primarily focus on static knowledge graphs, overlooking temporal dynamics critical for real-world knowledge evolution.

In the temporal domain, a few studies have begun applying LLMs to TKGs. TIMEBENCH~\cite{chu2024timebench} provides a benchmark explicitly assessing LLMs' temporal reasoning abilities. Approaches such as RoG~\cite{luoreasoning} utilize sequential fine-tuning on historical data to improve temporal coherence, while PPT~\cite{xu2023pre}, GPT-NeoX~\cite{lee2023temporal}, and Mixtral-8x7B-CoH~\cite{xia2024chain} employ prompting strategies tailored for temporal contexts. {Despite these advancements, a fundamental limitation persists in existing RAG and re-ranking paradigms \cite{wei2023kicgpt, wang2024kc}. These methods typically treat the TKG as a passive external memory, utilizing retrieved subgraphs merely as context to ground the LLM. Consequently, they are susceptible to propagating retrieval noise and lack the capability to rectify structural errors. distinct from these approaches, our work introduces an active structural refinement paradigm. Instead of simply consuming retrieved paths, we position the LLM as a logic validator to explicitly edit and repair the reasoning chains, thereby addressing the temporal hallucinations and logical gaps that passive retrieval methods cannot resolve.}

\section{Problem formulation}

 \textit{Temporal Knowledge Graph} is a structured representation of knowledge that captures the temporal evolution of entities and their relations over time. Formally, a TKG is defined as a quadruple:
$\mathcal{G} = (\mathcal{E}, \mathcal{R}, \mathcal{T}, \mathcal{Q})$,
where $\mathcal{E}$ is the set of entities, $\mathcal{R}$ is the set of relations, $\mathcal{T}$ is the set of discrete time points, $\mathcal{Q} \subseteq \mathcal{E} \times \mathcal{R} \times \mathcal{E} \times \mathcal{T}$ is the set of temporal facts, represented as quadruples $(s, r, o, t)$, where $s \in \mathcal{E}$ is the subject entity, $o \in \mathcal{E}$ is the object entity, $r \in \mathcal{R}$ is the relation and $t \in \mathcal{T}$ is the timestamp at which the fact holds.
Unlike static knowledge graphs, TKGs incorporate a temporal dimension, enabling modeling of dynamic relations and supporting reasoning over evolving knowledge.

\textit{Temporal Knowledge Graph reasoning} is the task of predicting missing entities in future facts given historical observations. Given a TKG with a set of observed quadruples $\mathcal{Q}_{\text{train}} = \{(s_i, r_i, o_i, t_i)\}$, the goal is to infer missing information in future queries of the form:
$\mathcal{Q}_{\text{query}} = \{(s_q, r_q, ?, t_q) \mid t_q > t_i\}$,
Formally, TKGR aims to learn a function:
$f: (\mathcal{E}, \mathcal{R}, \mathcal{T}) \rightarrow \mathcal{E}$,
which predicts the most probable object entity $\hat{o} \in \mathcal{E}$ for a given query $(s_q, r_q, ?, t_q)$, 
based on historical patterns and temporal dependencies.
An effective reasoning model should not only maximize predictive accuracy, but also ensure temporal consistency and interpretability, 
allowing trustworthy reasoning over evolving knowledge graphs.

\section{Methodology}

\begin{figure}[t] 
\centering
\includegraphics[width=1.0\textwidth]{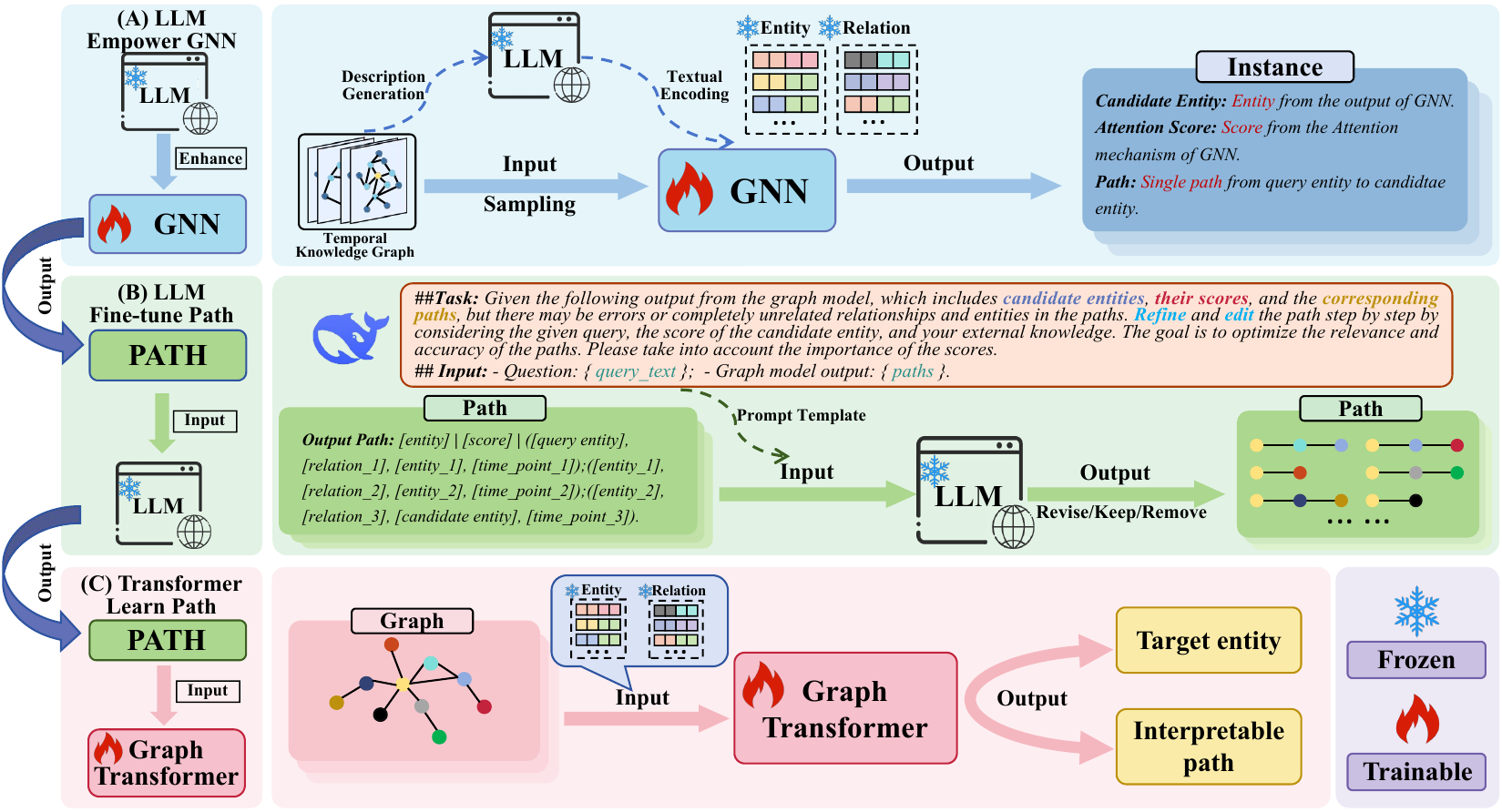}
\caption{The architecture of IGETR. The model consists of three key stages: (A) \textbf{LLM-empowered GNN reasoning}, where a sampling-based GNN extracts candidate entities and reasoning paths from the Temporal Knowledge Graph (TKG) with attention-based scoring; (B) \textbf{LLM-guided path refinement}, where the extracted paths are iteratively revised and optimized to improve logical consistency and relevance; and (C) \textbf{Graph Transformer-based learning}, which processes the refined paths to enhance interpretability and structural coherence. By integrating GNNs for structured data representation and LLMs for contextual reasoning, IGETR ensures both accurate and explainable TKGR. }
\label{fig:architecture}
\end{figure} 

The IGETR reasoning model is a novel hybrid framework designed to effectively integrate structural reasoning and contextual understanding by combining GNN with LLM. {This is accomplished through a three-stage pipeline, the architecture of which is depicted in Figure~\ref{fig:architecture}.} Initially, entity and relation embeddings are semantically enriched by leveraging textual encodings from an LLM, which provides the graph neural model with valuable external context. {Following this, we employ a candidate-oriented temporal GNN, which is chosen for its ability to focus reasoning on the most relevant local structures, to extract candidate paths from the TKG and produce initial reasoning chains grounded in structural and semantic evidence (\textit{Section A})}. Subsequently, {to overcome the data-bound limitations of the GNN and infuse external knowledge,} these preliminary paths undergo refinement via a prompt-based path optimization strategy employing an LLM, effectively addressing logical inconsistencies and irrelevant information to ensure path coherence and interpretability (\textit{Section B}). Lastly, a Graph Transformer-based aggregation module synthesizes multiple optimized reasoning paths, explicitly integrating temporal context and structural dependencies to produce reliable and interpretable final predictions (\textit{Section C}). Collectively, this structured integration of GNNs and LLMs within the IGETR framework significantly enhances the accuracy, logical consistency, and transparency of temporal knowledge graph reasoning.

\subsection{Stage I: Semantic-Aware Temporal Representation Learning}
In this section, we first describe how IGETR integrates a LLM with a temporal GNN model to support the extraction of the most relevant reasoning paths from the TKG. Initially, the entity and relation embeddings are enriched via textual encoding provided by the LLM, ensuring that the graph model captures semantic context from external knowledge. Subsequently, these context-enhanced embeddings guide a temporal GNN equipped with attention edge sampling, selectively identifying temporally coherent and semantically relevant candidate paths. The attention mechanisms in the GNN, combined with LLM-enhanced embeddings, effectively prioritize chronologically proximal and contextually consistent edges while reducing irrelevant connections. This collaborative approach supports the extraction of high-quality and interpretable candidate paths, forming a solid foundation for subsequent path refinement and aggregation.

\subsubsection{Semantic Embedding Initialization}

\begin{figure} 
	\centering
	\includegraphics[scale=0.6]{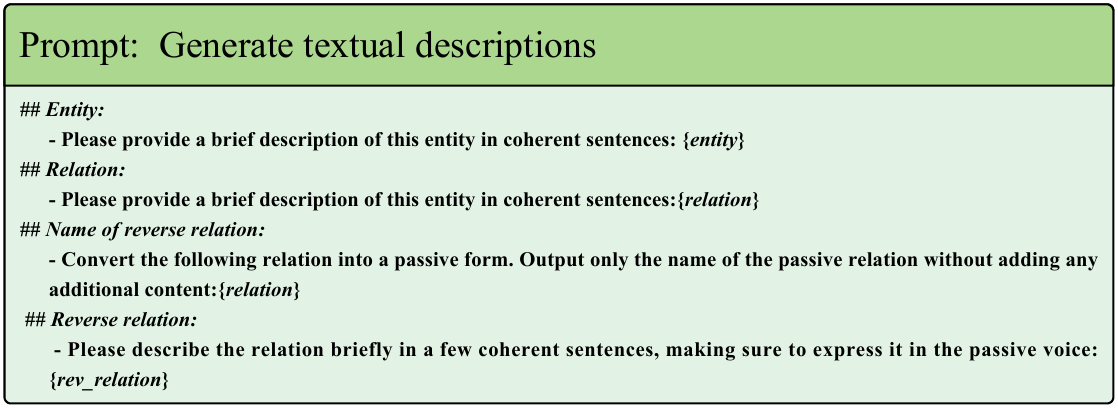}
	\caption{Prompt design for semantic embedding initialization. The prompt template guides the LLM to generate textual descriptions for entities and relations, as well as passive forms for reverse relations, enhancing semantic understanding for subsequent embedding and reasoning steps.}
	\label{fig:prompt1}
\end{figure}
We begin by generating concise textual descriptions for each entity $e$ and relation $r$ (including inverse relations) within the TKG using a LLM. Specifically, given the entities and relations, the LLM generates descriptive texts via a structured prompt, as illustrated in Figure \ref{fig:prompt1}. Formally, this process can be described as:
\begin{equation}
    Desc(e), Desc(r) = \mathrm{LLM_{gen}}(prompt(e,r)).
\end{equation}

The generated textual descriptions \( Desc(e) \) and \( Desc(r) \) are then encoded using an LLM textual encoding model to produce semantic embeddings for entities and relations. For each entity \( e \in \mathcal{E} \) and relation \( r \in \mathcal{R} \), we obtain initial semantic embeddings as follows:
\begin{equation}
    \textbf{H}_{gen(e)} = \mathrm{LLM}_{enc}(Desc(e)),\quad \textbf{H}_{gen(r)} = \mathrm{LLM}_{enc}(Desc(r)),
\end{equation}
where \( \textbf{H}_{gen(e)} \in \mathbb{R}^{|\mathcal{E}| \times d_w} \) and \( \textbf{H}_{gen(r)} \in \mathbb{R}^{|\mathcal{R}|\times d_w} \), and \( d_{w} \) denotes the embedding dimension of the encoder output.

To align these embeddings with the input dimension required by the subsequent GNN model, we introduce two multi-layer perceptrons (MLPs) to project the embeddings into the appropriate space:
\begin{equation}
    \textbf{h}_{e} = \mathrm{MLP}_{e}(\textbf{H}_{gen(e)}),\quad \textbf{h}_{r} = \mathrm{MLP}_{r}(\textbf{H}_{gen(r)}),
\end{equation}
where \( \textbf{h}_e \) and \( \textbf{h}_{r} \) denote the final semantic embeddings of entities and relations, respectively. These enriched embeddings help the GNN module better capture semantic nuances and contextual relations within the temporal graph structure, thereby enhancing the quality and interpretability of extracted reasoning paths.

\subsubsection{Temporal GNN with Attention-guided Sampling}
{In our framework, we employ a specific type of temporal GNN architecture derived from our previous work, CaORG~\cite{fan2024flow}. This model is a subgraph-based temporal GNN designed for forecasting tasks. Unlike conventional temporal GNNs that typically operate on fixed neighborhoods or the entire graph, CaORG dynamically constructs a unique candidate-oriented relational graph for each query-candidate pair. This approach allows the model to focus reasoning on the most relevant local structures and logical paths connecting a query to a potential answer.}

{To effectively embed temporal information within this architecture, we first encode time using random Fourier features, following a similar methodology to TGAT~\cite{xu2020inductive}}, and combine these temporal encodings with the relation embeddings:
$\textbf{h}_{r,t}=\textbf{w}_{r,t}\left[ \textbf{h}_{r} \| \Phi(t) \right]$,
where $\textbf{w}_{r,t}$ denotes a learnable transformation, and $\textbf{h}_{r,t}$ represents the temporal embedding for relation $r$ at time $t$.

Subsequently, the temporal-aware embeddings are used in our attention-enhanced GNN to propagate entity information along temporally coherent paths. Specifically, at each layer $l$, the representation of an entity $e_o$ is updated by aggregating messages from neighboring entities $e_s$:
\begin{equation}
    \textbf{h}_{o}^{l}=\sigma\left(\textbf{W}_{h}^{l}\cdot AGGREGATE_{l}\left\{\alpha_{s,o}^{l}\cdot\left(\textbf{h}_{s}^{l-1}+\textbf{h}_{r,t}^{l-1}\right),\forall{e_s}\in {\mathcal{N}_{\left(o\right)}^{l-1}}\right\}\right),
\end{equation}
where $\sigma(\cdot)$ denotes an activation function (e.g., ReLU), and $\textbf{h}_o^l$ represents the embedding of entity $e_o$ at the $l$-th layer. At layer $l-1$, $\textbf{h}_{s}^{l-1}$ denotes the embedding of a neighboring entity $e_s$. Given a temporal query quadruple $(e_q,r_q,?,t_q)$, we initialize entity embeddings as $\textbf{h}_{e_s}^0=\textbf{h}_{e_q}$. Relation embedding $\textbf{h}_{r,t}$ encodes the semantic and temporal information of the relation $r$ between entities $e_s$ and $e_o$ at timestamp $t$. To construct the local neighborhood for each entity $e_o$, neighbor entities $\mathcal{N}_{o}^{l-1}$ are sampled using an attention-guided approach, enhancing relevance and interpretability.

The aggregation function integrates neighborhood information guided by attention scores $\alpha_{s,o}^{l}$. Specifically, at each layer, the message passed from neighbor entity $e_s$ to target entity $e_o$ is computed by combining entity embeddings $\textbf{h}_{s}^{l-1}$, relation embeddings $\textbf{h}_{r,t}$, and the query relation embedding $\textbf{h}_{r_q,t_q}^{l}$. The attention mechanism assigns higher weights to more relevant and temporally coherent neighbors, thereby effectively capturing structural and semantic information essential for accurate TKGR:
\begin{equation}
    \alpha_{s,o}^{l}=\sigma\left(\left(\textbf{w}_{\alpha}^{l}\right)^{T}\cdot\delta\left(\textbf{W}_{\alpha}\cdot CONCAT\left(\textbf{h}_{s}^{l-1},\textbf{h}_{r,t}^{l},\textbf{h}_{r_{q},t_{q}}^{l}\right)\right)\right),
\label{attention}
\end{equation}
where $\delta(\cdot)$ denotes the ReLU activation function, $\textbf{h}_{r_q,t_q}^{l}$ is the embedding of the query relation at query time $t_q$, $\textbf{W}_{\alpha}^{l}$ and $\textbf{w}_{\alpha}^{l}$ are learnable parameters, and $\|$ denotes concatenation. The attention mechanism ensures that entities and relations most relevant to the temporal query are effectively prioritized, producing informative reasoning paths.

Finally, given candidate entities extracted through this GNN-based reasoning process, 
we calculate their scores to assess the likelihood of each candidate entity $e_i$ being the correct answer:
$a_{e_i}(q)=\textbf{w}_a^{T}\cdot \textbf{h}^{L}_{e_{i}}$,
where $\textbf{w}_a$ is a learnable vector and $\textbf{h}_{e_i}^L$ is the final-layer embedding of candidate entity $e_i$. The entire network is optimized via a cross-entropy loss defined as:
\begin{equation}
    \mathcal{L}=-\sum_{(e_q,r_q,e_a,t_q)\in\mathcal{F}_{train}}\log\frac{\exp(a_{e_a}(q))}{\sum_{e_i\in\mathcal{E}}\exp(a_{e_i}(q))},
\end{equation}
where $\mathcal{F}_{train}$ denotes the set of positive quadruples in the training dataset.

{The training process is fully supervised, with the ground-truth entity $e_a$ serving as the direct supervision signal. By minimizing this cross-entropy loss, the network learns to identify coherent reasoning paths as an emergent property. Temporal coherence is enforced by the model's design, as path construction is restricted to historical facts. Structural coherence, in turn, is learned implicitly by the attention mechanism (Eq.\ref{attention}). The backpropagation process adjusts the attention weights to prioritize paths that are most predictive of the correct answer, effectively training the GNN to recognize and up-weight structurally and semantically relevant evidence. This allows the model to extract accurate and interpretable reasoning paths for subsequent refinement.}

\subsection{Stage II: Refining Paths by Prompt Optimization}

In this section,  we describe how to extract relevant reasoning paths from the temporal GNN and refine them using a LLM. Although the paths obtained from the temporal GNN already encode temporal coherence and structural information, they may contain redundant, incorrect, or semantically inconsistent relations. To address these limitations, we employ a prompt-based approach, leveraging the LLM's external knowledge and contextual reasoning capabilities to iteratively revise, keep, or remove elements within the extracted paths. By doing so, we substantially enhance the accuracy, logical consistency, and interpretability of the reasoning chains, ensuring their suitability for trustworthy reasoning in downstream tasks.

In this section, we describe how to extract relevant reasoning paths from the temporal GNN and refine them using a LLM. Although the paths obtained from the temporal GNN already encode temporal coherence and structural information, they may contain redundant, incorrect, or semantically inconsistent relations. {To address these limitations, we employ a prompt-based editing mechanism that fundamentally differs from passive re-ranking strategies. Instead of merely selecting from fixed candidates, our approach leverages the LLM's external knowledge to actively \textit{revise, keep, or remove} elements within the extracted paths. By doing so, we substantially enhance the accuracy, logical consistency, and interpretability of the reasoning chains, ensuring their suitability for trustworthy reasoning in downstream tasks.}

\subsubsection{Candidate Path Construction}
{To extract valuable inference chains for subsequent refinement by the LLM, we employ a deterministic ranking and pruning strategy. First, given the final entity prediction scores, we select the top-$K$ candidate entities $\{e_i\}_{i=1}^{K}$ with the highest probabilities as reference answers}:
\begin{equation}
    \mathcal{E}_{c} = \mathrm{top\text{-}K}\left(\{a_{e_i}(q) \mid e_i \in \mathcal{E}\}\right).
\end{equation}
Next, for each candidate entity $e_i \in \mathcal{E}_{c}$, we backtrack iteratively using a greedy selection strategy. At each step \(l\), we select the neighbor entity $e_{l-1}^{*}$ that maximizes the attention weight leading to the current entity $e_l$, formally expressed as:
\begin{equation}
    e_{l-1}^{*} = \arg\max_{e_{l-1}\in \mathcal{N}_{e_l}} \alpha_{e_{l-1}, e_l}^{l}.
\end{equation}
Starting from candidate answer $e_i$, we iteratively backtrack through entities according to this greedy strategy, forming a reasoning path until the query entity $e_q$ is reached:
\begin{equation}
\mathcal{P}_i(e_q, e_i) = (e_q, r_1, e_1, t_1) \rightarrow (e_1, r_2, e_2, t_2) \rightarrow \dots \rightarrow (e_{L-1}, r_L, e_i, t_L),
\end{equation}
where $L$ is the maximum length of the reasoning chain, 
and each step $(e_{l-1}$, 
$r_l$, $e_l$, $t_l)$ 
corresponds to the edge with the highest attention weight at layer $l$. This process produces a high-quality and interpretable inference path linking the query entity $e_q$ to the candidate answer $e_i$, thus facilitating subsequent logical refinement by the LLM.

\subsubsection{{Formalized Path Editing}}

\begin{figure}[t]
	\centering
	\includegraphics[scale=0.6]{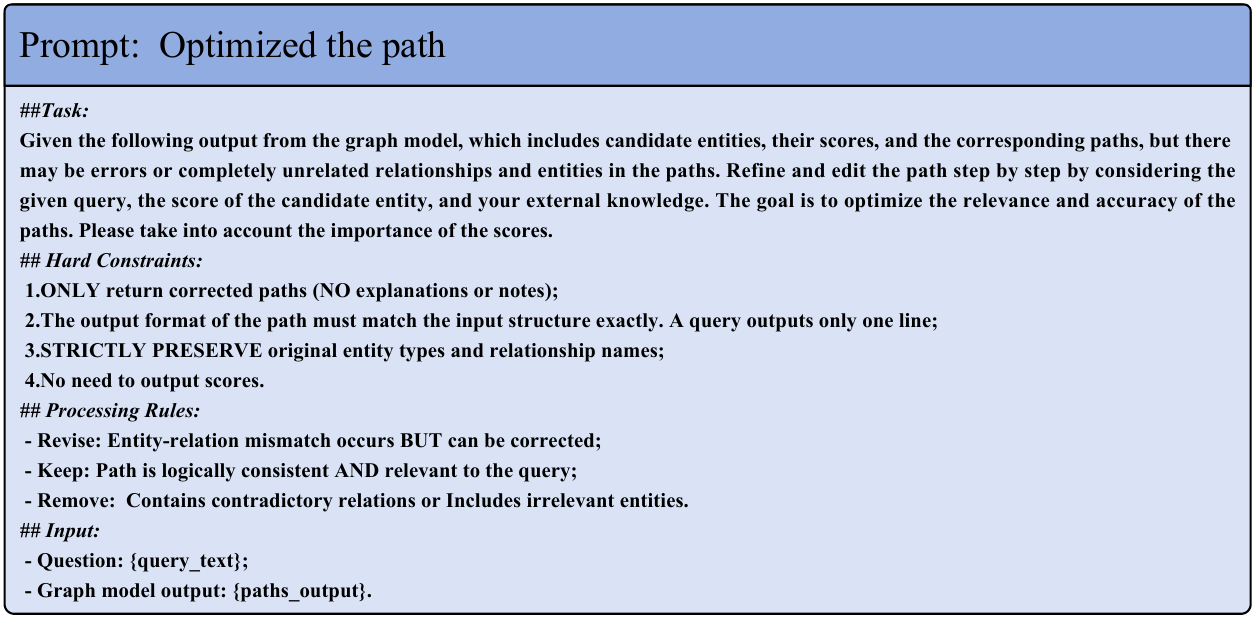}
	\caption{Prompt design for path optimization. The structured prompt clearly instructs the LLM to revise, keep, or remove segments of extracted inference paths according to specified logical constraints and processing rules, thus ensuring optimized relevance, accuracy, and interpretability of the reasoning paths.}
	\label{fig:prompt2}
\end{figure}

{To mitigate the risk of hallucinations and ensure logical rigor, we formalize the path refinement process as a constrained optimization task. Specifically, given a candidate path $\mathcal{P}_i = \{(e_{l-1}, r_l, e_l, t_l)\}_{l=1}^L$ extracted by the GNN, the LLM functions as a transformation operator $LLM_{edit}$ that maps $\mathcal{P}_i$ to a refined path $\tilde{\mathcal{P}}_i$.}

{The optimization is driven by the prompt $\psi(\mathcal{P}_i)$, constructed as shown in Figure \ref{fig:prompt2}. Crucially, we directly utilize the dataset-specific discrete time identifiers within the prompt, rather than incorporating absolute real-world timestamps. This design prevents the model from retrieving parametric historical knowledge associated with specific dates, thereby forcing it to rely solely on the logical consistency within the provided context.}

{The LLM generates an operation sequence applied to the path elements. We formally define the action space of the operators and their associated constraints $\mathcal{C}$ as follows:}
\begin{itemize}[leftmargin=*]
    \item \textbf{Keep:} {The path segment is retained unchanged if it is logically coherent and semantically relevant to the query. Formally, the output segment remains $(\tilde{e}_l, \tilde{r}_l) = (e_l, r_l)$.}
    
    \item \textbf{Remove:} {The segment is discarded if it contains contradictory relations or irrelevant entities. This effectively prunes the invalid branch from the reasoning graph.}
    
    \item \textbf{Revise:} {The segment is modified to correct entity-relation mismatches. To ensure the revised step $(\tilde{e}_l, \tilde{r}_l)$ remains valid within the graph structure, we enforce strict hard constraints $\mathcal{C}$ on this operator. First, the revised entity and relation must belong to the existing graph vocabulary ($\tilde{e}_l \in \mathcal{E}$ and $\tilde{r}_l \in \mathcal{R}$). Second, we enforce type consistency, requiring that the semantic type of the revised entity must match the original, ensuring ontological validity. Finally, to enforce temporal causality, the timestamp must strictly adhere to the chronological order of the chain $t_{l-1} \le t_l \le t_{l+1}$.}
\end{itemize}

{Formally, the optimization process is represented as:
\begin{equation}
\tilde{\mathcal{P}}_i(e_q, e_i) = LLM_{edit}
\left(\psi\left(\mathcal{P}_i\right), \mathcal{C}\right),
\end{equation}
where $\mathcal{C}$ denotes the set of constraints defined above. This approach ensures strict adherence to the original entity types while leveraging the LLM's semantic reasoning to rectify logical inconsistencies.}

\subsection{Stage III: Aggregating Paths through Temporal Graph Transformer}

To effectively aggregate and refine optimized reasoning paths obtained from the LLM-enhanced temporal GNN, we leverage a Graph Transformer model inspired by Ruleformer~\cite{xu2022ruleformer}, which we extend to explicitly incorporate temporal information. Our primary objective is to aggregate multiple refined paths coherently, maintaining structural integrity and temporal consistency to facilitate accurate and interpretable predictions.

{Given an optimized reasoning path of length $L$, we first construct the input sequence embedding matrix $\mathbf{X}$. Unlike the original Ruleformer which focuses on static rules, we fuse temporal context into each step $l$:}
\begin{equation}
\mathbf{x}_{l} = \mathbf{h}_{e_l} + \mathbf{h}_{r_l} + \Phi(t_l), \quad l \in [1, L],
\end{equation}
where $\mathbf{h}_{e_l}, \mathbf{h}_{r_l}$ denote entity and relation embeddings, and $\Phi(t_l)$ represents the temporal positional encoding parameterized by random Fourier features.

{We then apply a Transformer encoder to perform structure-aware aggregation. The core mechanism is a relational attention module that computes the semantic relevance between any two steps $u$ and $v$ in the path:}
\begin{equation}
a_{u,v} = \frac{1}{\sqrt{d_{k}}}
\left(\mathbf{h}_{v}^{\top}\mathbf{W}^{Q}\right)
\left(
\mathbf{h}_{u}\mathbf{W}^{K} + \sum_{r\in\mathcal{R}} k_{r,v,u}\mathbf{h}_{r}\mathbf{W}^{K'}
\right)^{\top},
\end{equation}
{where $\mathbf{W}^{Q}, \mathbf{W}^{K}, \mathbf{W}^{K'}$ are projection matrices. The term $\sum k_{r,v,u}\mathbf{h}_{r}$ explicitly injects the edge information between nodes into the attention calculation, allowing the model to weigh evidence based on structural connectivity.}

{Finally, the Transformer outputs an aggregated path representation $\mathbf{h} = \text{Transformer}(\mathbf{X})$. This vector is decoded to predict the validity score for the candidate entity $e_i$:
\begin{equation}
\mathbf{o}_i = \mathbf{W}_o \mathbf{h}, \quad \hat{e} = \arg\max_{e \in \mathcal{E}}(\mathbf{o}_i).
\end{equation}
Here, $\mathbf{W}_o$ is the learnable weight matrix for the output layer, mapping the path representation $\mathbf{h}$ to the output scores $\mathbf{o}_i$. By explicitly incorporating temporal encoding $\Phi(t)$ into the Ruleformer-based architecture, this module ensures that the final prediction is grounded in a logically verified and temporally consistent evidence chain.}

\subsection{Technical Comparison}
\begin{table}[t]
\centering
\caption{Technical Comparison of Representative TKGR Models}
\resizebox{\textwidth}{!}{
\renewcommand\arraystretch{1.5}
\begin{tabular}{l|c|c|c|c}
\hline
Model             & Reasoning Strategy                   & Temporal Awareness & External Knowledge & Information Filter \\ \hline
TLogic            & Multi-hop Path                       & Yes \textcolor{green}{\ding{51}}                & No {\ding{55}}                  & No {\ding{55}}                 \\
TiRGN             & Multi-hop Graph                      & Yes \textcolor{green}{\ding{51}}                & No {\ding{55}}                  & No {\ding{55}}                 \\
CoH               & Direct Interaction                   & Limited            & Yes \textcolor{green}{\ding{51}}                & No {\ding{55}}                 \\
CoH + Graph Model & Multi-hop Graph + Direct Interaction & Partial            & Yes \textcolor{green}{\ding{51}}                & No {\ding{55}}                 \\
IGETR             & Multi-hop Graph + Direct Interaction & Yes \textcolor{green}{\ding{51}}                & Yes \textcolor{green}{\ding{51}}                & LLM-based Path Edit \textcolor{green}{\ding{51}}         \\ \hline
\end{tabular}
}
\label{tab:compare}
\end{table}

Table~\ref{tab:compare} presents a comparative analysis of representative TKGR models across four key dimensions: reasoning strategy, temporal awareness, integration of external knowledge, and information filter. Traditional models such as TLogic~\cite{liu2022tlogic} and TiRGN~\cite{li2022tirgn} perform reasoning purely over graph structures using multi-hop paths or subgraphs, but they neither integrate external knowledge nor support refine the reasoning paths. CoH~\cite{luo2024chain}, on the other hand, enables direct interaction with queries using LLMs and leverages external knowledge, yet lacks temporal structure awareness and controllable interpretable reasoning. A hybrid approach combining CoH with graph-based models attempts to incorporate both structural and semantic reasoning. However, it still falls short of controlling the reasoning process, offering no means to review or correct generated paths. In contrast, our proposed IGETR framework stands out by supporting both multi-hop graph traversal and LLM-based direct interaction. Integrate external knowledge through a path editing module and ensure strong temporal awareness grounded in structured evidence. This design allows IGETR to deliver more interpretable, controllable, and temporally coherent reasoning compared to all baselines.

\section{Experiment}
In this section, we systematically evaluate the effectiveness and interpretability of the proposed IGETR framework through comprehensive experiments. Specifically, we first introduce the datasets employed in our evaluation, providing essential statistical details and characteristics relevant to temporal reasoning tasks. Next, we present the baseline models selected for comparative analysis, include recent representative models from both GNN-based and LLM-based temporal knowledge graph reasoning methods. Subsequently, we present our main experimental results, emphasizing IGETR's improvements in predictive accuracy compared to these baselines. Additionally, we conduct ablation studies to analyze the contributions of main modules within IGETR, and the measures employed to prevent LLM from directly retrieving internal historical data during predictions. Finally, a case study is provided to qualitatively demonstrate the validity coherence of the LLM-based path editing mechanism, highlighting its capability in correcting and refining reasoning paths.

\subsection{Datasets}

\begin{table}[t]
\centering
\caption{Statistics of the datasets}
\tiny  
\renewcommand\arraystretch{1.5}  
\resizebox{\textwidth}{!}{  
\begin{tabular}{lcccccc}
\hline
\textbf{Datasets}  & \textbf{$\mathcal{E}$} & \textbf{$\mathcal{R}$} & \textbf{Train} & \textbf{Valid} & \textbf{Test} & \textbf{Time gap} \\ \hline
\textbf{ICEWS14}   & 6869       & 230        & 74845          & 8514           & 7371          & day               \\
\textbf{ICEWS18}   & 23033      & 256        & 373018         & 45995          & 49545         & day               \\
\textbf{ICEWS0515} & 10094      & 251        & 368868         & 46302          & 46159         & day               \\ \hline
\end{tabular}
}
\label{tab:datasets}
\end{table}

We evaluated our proposed model using four publicly available TKG datasets: ICEWS14, ICEWS05-15 and ICEWS18. The Integrated Crisis Early Warning System (ICEWS) dataset~\cite{DVN/28075_2015} records political events worldwide along with associated timestamps. Specifically, ICEWS14, ICEWS18, and ICEWS05-15 are subsets of ICEWS events collected within the years 2014, 2018, and between 2005 and 2015, respectively.

Following the standard evaluation protocol utilized in previous works such as RE-NET~\cite{jin2020recurrent}, we partitioned each dataset chronologically into three subsets: training, validation, and testing, ensuring temporal order $t_{train}<t_{valid}<t_{test}$. Detailed statistical information for each dataset, including the number of entities, relations, temporal facts, and timestamp granularity, is summarized in Table~\ref{tab:datasets}.

\subsection{Baseline}

The models selected for comparative analysis fall primarily into two categories: graph-based methods and LLM-based approaches. Within the category of graph-based methods, we evaluated several representative models: RE-NET~\cite{jin2020recurrent}, which employs a recurrent event encoder to effectively capture historical temporal patterns; RE-GCN~\cite{li2021temporal}, extending RE-NET with relational graph convolutional networks to enhance neighborhood aggregation; TiRGN~\cite{li2022tirgn}, leveraging temporal wandering mechanisms to explicitly model temporal dynamics; xERTE~\cite{han2021learning}, which dynamically prunes irrelevant information through query-related subgraph extraction; CaORG~\cite{fan2024flow}, which utilizes candidate-oriented relational graph reasoning to explicitly direct message propagation toward candidate entities, thereby enhancing the relevance and effectiveness of reasoning paths; TANGO~\cite{han2021learning}, which utilizes temporal-aware neighborhood sampling to improve reasoning accuracy; and HGLS~\cite{zhang2023learning}, which integrates hierarchical graph structures to capture comprehensive global and local historical patterns.

Regarding LLM-based methods, we compare our model against GenTKG~\cite{liao2024gentkg}, which leverages temporal logic interactions tailored for large language models, and advanced methods such as GPT-NeoX~\cite{lee2023temporal}, Llama-2-7b-CoH and Vicuna-7b-CoH~\cite{luo2024chain}, as well as Mixtral-8x7B-CoH~\cite{xia2024chain}, all of which utilize sophisticated prompt-based and in-context learning strategies to infer temporal events. These comparisons comprehensively highlight our model's strengths in accuracy, interpretability, and temporal reasoning capability.

\subsection{Implementation Details}
In our experiments, entity and relation embeddings were initialized by employing GLM-4-Flash as the text-generation model $LLM_{gen}$, while text-embedding-v3 served as the textual encoding model $LLM_{enc}$. The Deepseek-v3 model was adopted as the path-editing $LLM_{edit}$. {All hyperparameters were tuned based on the model's performance on the validation set to prevent overfitting and ensure a fair evaluation. We employed a grid search over the following parameter ranges: learning rate in \{0.0005, 0.001, 0.003\}, batch size in \{32, 64, 128\}, reasoning path length $L$ in \{2, 3, 4\}, and embedding dimensions in \{128, 256\}. The best-performing configuration, with a learning rate of 0.001, a batch size of 64, and a path length of $ L=3$, was selected for the final evaluation. The model was trained using the Adam optimizer. The final results reported in this paper were obtained by running this optimal model on the unseen test set. For each query, we selected $K=30$ candidate paths for LLM refinement. This larger budget was a deliberate choice to ensure a diverse set of reasoning paths was considered for refinement, rather than only the top few highest-scoring ones.}

\subsection{Main result}

\begin{table*}[t]\large
\centering
\caption{Performance comparison on TKGR task. The best results among all models are in bold and the second best results among all models are underlines.}
\resizebox{\textwidth}{!}{
\renewcommand\arraystretch{1.5}

\begin{tabular}{lccccccccc}
\hline
\multicolumn{1}{c}{\multirow{2}{*}{\textbf{Model}}} & \multicolumn{3}{c}{\textbf{ICEWS14}}                                  & \multicolumn{3}{c}{\textbf{ICEWS0515}}                                & \multicolumn{3}{c}{\textbf{ICEWS18}}             \\ \cline{2-10} 
\multicolumn{1}{c}{}                                & Hits@1         & Hits@3         & Hits@10                             & Hits@1         & Hits@3         & Hits@10                             & Hits@1         & Hits@3         & Hits@10        \\ \hline
\multicolumn{1}{l|}{RE-NET}                         & 0.293          & 0.431          & \multicolumn{1}{c|}{0.575}          & 0.334          & 0.478          & \multicolumn{1}{c|}{0.611}          & 0.192          & 0.323          & 0.483          \\
\multicolumn{1}{l|}{RE-GCN}                         & 0.297          & 0.441          & \multicolumn{1}{c|}{0.586}          & 0.336          & 0.487          & \multicolumn{1}{c|}{0.658}          & 0.193          & 0.331          & 0.494          \\
\multicolumn{1}{l|}{xERTE}                          & 0.312          & 0.453          & \multicolumn{1}{c|}{0.570}          & 0.347          & 0.497          & \multicolumn{1}{c|}{0.633}          & 0.206          & 0.330          & 0.458          \\
\multicolumn{1}{l|}{TANGO}                          & 0.151          & 0.272          & \multicolumn{1}{c|}{0.431}          & 0.311          & 0.476          & \multicolumn{1}{c|}{0.622}          & 0.178          & 0.314          & 0.460          \\
\multicolumn{1}{l|}{CaORG}                          & 0.325          & 0.479          & \multicolumn{1}{c|}{0.612}          & 0.364          & {\ul 0.541}    & \multicolumn{1}{c|}{0.678}          & 0.208          & 0.335          & 0.479          \\
\multicolumn{1}{l|}{TLogic}                         & 0.322          & 0.470          & \multicolumn{1}{c|}{0.603}          & 0.345          & 0.525          & \multicolumn{1}{c|}{0.673}          & 0.205          & 0.339          & 0.484          \\
\multicolumn{1}{l|}{TiRGN}                          & 0.313          & 0.468          & \multicolumn{1}{c|}{0.612}          & 0.358          & 0.535          & \multicolumn{1}{c|}{0.690}          & 0.202          & 0.339          & 0.484          \\
\multicolumn{1}{l|}{HGLS}                           & {\ul 0.349}    & {\ul 0.480}    & \multicolumn{1}{c|}{\textbf{0.688}} & 0.351          & 0.521          & \multicolumn{1}{c|}{0.673}          & 0.192          & 0.323          & 0.494          \\ \hline
\multicolumn{1}{l|}{GenTKG}                         & {\ul 0.349}    & 0.473          & \multicolumn{1}{c|}{0.619}          & 0.360          & 0.525          & \multicolumn{1}{c|}{0.687}          & 0.215          & \textbf{0.366} & 0.496          \\
\multicolumn{1}{l|}{GPT-NeoX-20B-ICL}               & 0.295          & 0.406          & \multicolumn{1}{c|}{0.475}          & 0.360          & 0.497          & \multicolumn{1}{c|}{0.586}          & 0.177          & 0.290          & 0.385          \\
\multicolumn{1}{l|}{Llama-2-7b-ICL}                 & 0.275          & 0.391          & \multicolumn{1}{c|}{0.453}          & 0.353          & 0.490          & \multicolumn{1}{c|}{0.563}          & 0.177          & 0.295          & 0.364          \\
\multicolumn{1}{l|}{Vicuna-7b-ICL}                  & 0.270          & 0.386          & \multicolumn{1}{c|}{0.453}          & 0.347          & 0.483          & \multicolumn{1}{c|}{0.563}          & 0.172          & 0.288          & 0.364          \\
\multicolumn{1}{l|}{Llama-2-7b-COH}                 & 0.338          & 0.462          & \multicolumn{1}{c|}{0.587}          & 0.370          & 0.531          & \multicolumn{1}{c|}{{\ul 0.699}}    & {\ul 0.219} & 0.361          & 0.520          \\
\multicolumn{1}{l|}{Vicuna-7b-COH}                  & 0.315          & 0.445          & \multicolumn{1}{c|}{0.648}          & {\ul 0.372}    & 0.531          & \multicolumn{1}{c|}{\textbf{0.701}} & 0.206          & 0.344          & \textbf{0.531} \\ \hline
\multicolumn{1}{l|}{IGETR}                           & \textbf{0.357} & \textbf{0.519} & \multicolumn{1}{c|}{{\ul 0.652}}    & \textbf{0.393} & \textbf{0.552} & \multicolumn{1}{c|}{0.683}          & \textbf{0.225}              & {\ul 0.357}              & {\ul 0.530}              \\ \hline
\end{tabular}%
}

\label{tab:result}
\end{table*}

The experimental results comparing our proposed IGETR model with baseline methods on the ICEWS14, ICEWS05-15 and ICEWS18 datasets are presented in Table~\ref{tab:result}. IGETR consistently demonstrates superior performance across multiple metrics, significantly outperforming all baseline approaches. Specifically, on the ICEWS14 dataset, IGETR achieves the highest Hits@1 and Hits@3 scores, with improvements of 2.3\% and 8.1\%, respectively, compared to the second-best performing model. On ICEWS05-15, our method again shows strong performance, achieving improvements of 5.6\% in Hits@1 and 2.0\% in Hits@3 compared to the strongest baseline model. These results highlight the advantages of our proposed integration of graph structural reasoning and LLM-based semantic refinement. However, it is noteworthy that our approach edits only a fixed number of paths due to constraints limiting from the sampling limitations inherent in the graph model and the input capacity of the large language model. Consequently, these practical restrictions limit our model's ability to fully leverage available information, particularly affecting performance metrics such as Hits@10 or on datasets with denser historical interactions, exemplified by ICEWS18.

Compared to graph-based baseline models (such as RE-NET, RE-GCN, and CaORG), our IGETR model benefits significantly from leveraging the semantic enhancement provided by large language models. {This goes beyond simple semantic enrichment; our novel LLM-mediated path editing actively refines the reasoning chains, using external knowledge to correct logical inconsistencies that purely data-bound models cannot resolve.} While traditional graph-based methods rely solely on structural patterns derived from historical data, IGETR's approach leverages additional semantic context, thereby capturing more nuanced and contextually relevant relations. Conversely, when compared with purely LLM-based methods (such as GPT-NeoX-20B-ICL, Vicuna-7b, and Llama-2-7b variants), our model demonstrates greater accuracy due to the explicit structural guidance from the graph neural network, effectively reducing the "hallucination" issues commonly encountered by LLMs when reasoning purely based on textual prompts. {The key insight here is that the GNN provides a high-quality, verifiable scaffold of path candidates grounded in actual data, which constrains the LLM and ensures its powerful reasoning capabilities are applied to a factually sound foundation.} The complementary strengths of graph-based reasoning and LLM-based semantic reasoning are thus effectively harnessed in our framework, yielding improved accuracy and interpretability.

In summary, the performance improvements achieved by IGETR across the evaluated datasets validate the effectiveness of combining structural knowledge from GNNs and semantic context from LLMs, clearly distinguishing our model from conventional single-method approaches.

\subsection{Ablation Study}
In this section, we conduct comprehensive ablation studies to investigate critical aspects of our proposed IGETR model. Specifically, we first evaluate the effectiveness of the LLM-based path editing module, demonstrating how LLM-guided path refinement contributes to the accuracy. Subsequently, we design additional experiments to verify that the performance improvements from LLM-based path editing originate from enhanced logical reasoning rather than merely retrieving internal historical data. This analysis further confirms that our model's interpretability and robustness stem from genuine reasoning improvements enabled by the proposed LLM-empowered editing approach.

\subsubsection{The analysis of LLM edit module}
\begin{figure}
\centering
\subfloat[ICEWS14]{\includegraphics[width=.40\linewidth]{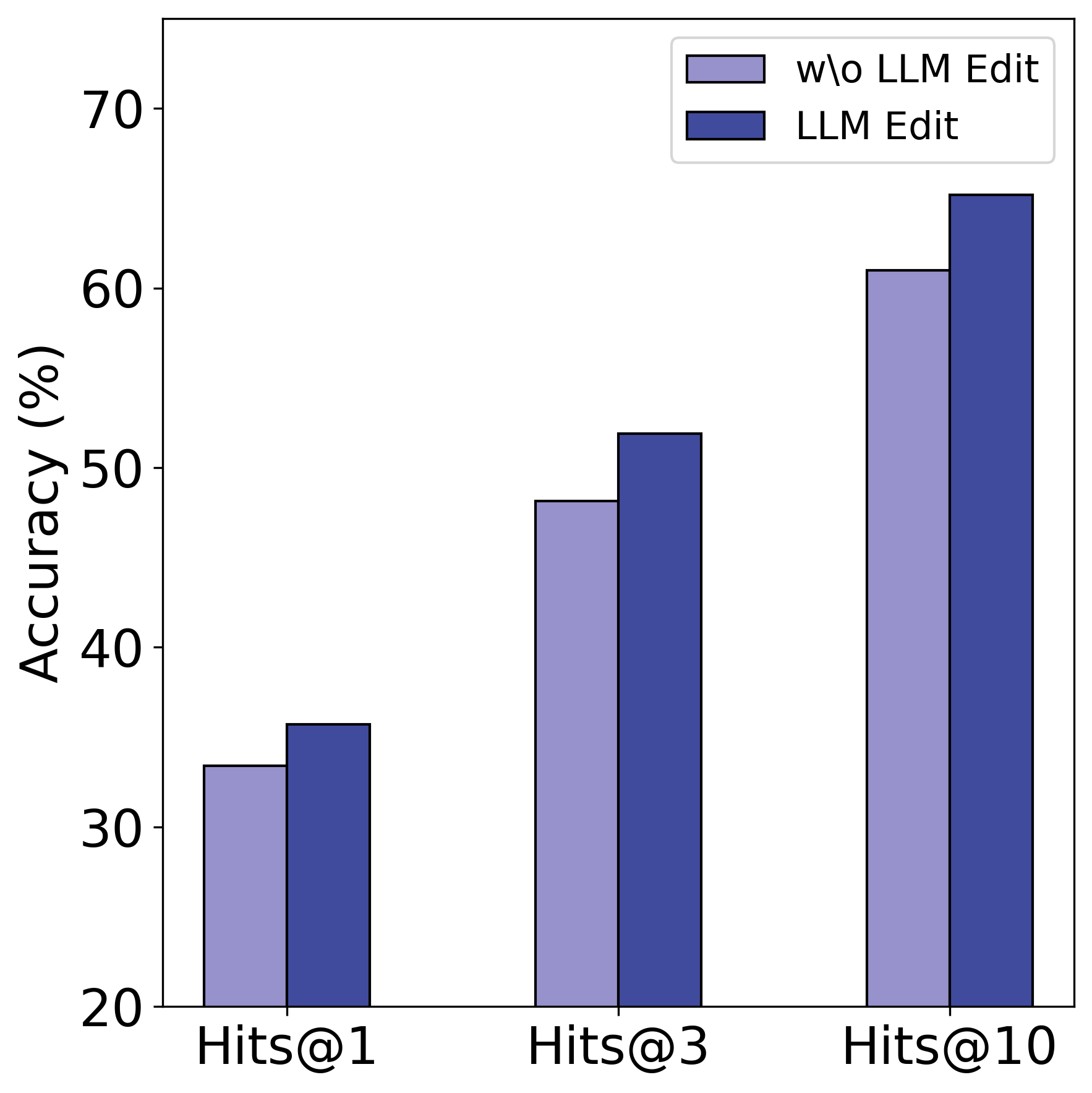}}\hspace{1pt}
\subfloat[ICEWS0515]{\includegraphics[width=.40\linewidth]{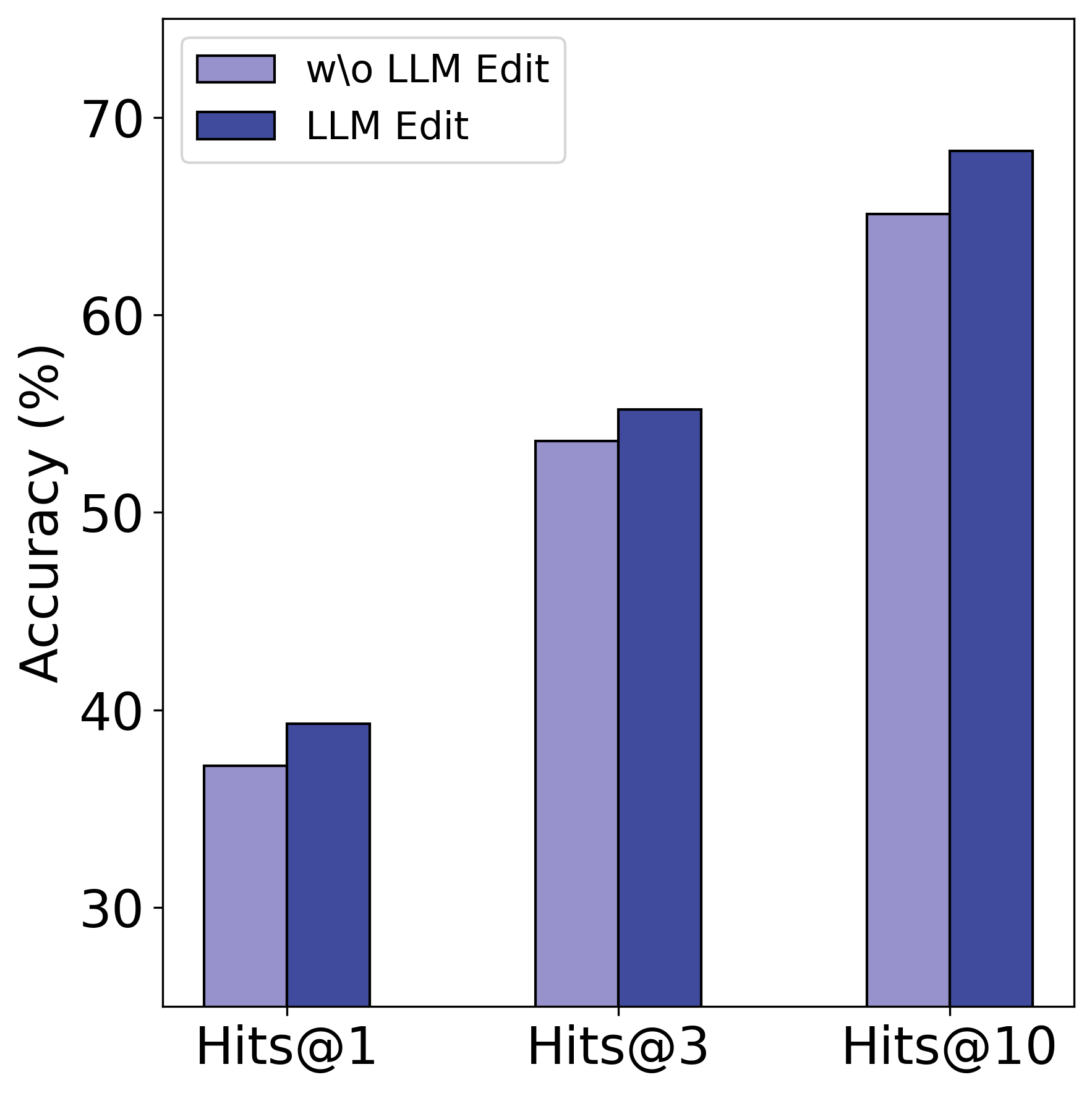}}
\caption{We conducted experiments on the ICEWS14 and ICEWS0515 datasets to demonstrate the effectiveness of the LLM-based path editing module. The results indicate a performance improvement across evaluation metrics  when incorporating the LLM-based path editing.}
\label{fig:ab_edit}
\end{figure}

To validate the contribution of the LLM-based path editing module, we conduct experiments comparing the performance of the complete IGETR framework with a variant from which the LLM path editing component has been removed. {We hypothesize that the LLM editing step is critical for refining inference paths and should therefore lead to a measurable improvement in prediction accuracy. To visualize this effect, Figure~\ref{fig:ab_edit} presents a direct comparison of the Hits@1, Hits@3, and Hits@10 metrics for both configurations on the ICEWS14 and ICEWS05-15 datasets. As the results clearly show, there is a noticeable drop in performance metrics after removing the LLM-based path editing step, highlighting the module's significant role in enhancing the semantic coherence and accuracy of the reasoning paths.}

\subsubsection{The analysis of LLM retrieve historical data}
\begin{table*}[t]\large
\centering
\caption{Experimental results of directly querying large language models using absolute timestamps.}
\resizebox{\textwidth}{!}{
\renewcommand\arraystretch{1.5}
\begin{tabular}{lccccccccc}
\hline 
\multicolumn{1}{c}{\multirow{2}{*}{\textbf{Model}}} & \multicolumn{3}{c}{\textbf{ICEWS14}}         & \multicolumn{3}{c}{\textbf{ICEWS0515}}       & \multicolumn{3}{c}{\textbf{ICEWS18}} \\ \cline{2-10} 
\multicolumn{1}{c}{}                                & Hits@1 & Hits@3 & Hits@10                    & Hits@1 & Hits@3 & Hits@10                    & Hits@1     & Hits@3     & Hits@10    \\ \hline
\multicolumn{1}{l|}{Deepseek-V3}                    & 0.052  & 0.123  & \multicolumn{1}{c|}{0.214} & 0.046  & 0.127  & \multicolumn{1}{c|}{0.223} & 0.053      & 0.115      & 0.196      \\
\multicolumn{1}{l|}{IGETR}                          & 0.357  & 0.519  & \multicolumn{1}{c|}{0.652} & 0.393  & 0.552  & \multicolumn{1}{c|}{0.683} & 0.225      & 0.357      & 0.530      \\ \hline
\end{tabular}
}

\label{tab:ab1}
\end{table*}

LLMs inherently possess vast amounts of pre-existing data, raising concerns that datasets used for experiments might overlap with the LLM training data. Consequently, there exists a potential risk that the LLM could directly retrieve historical records rather than perform genuine logical reasoning. To mitigate this issue, we deliberately replace the real-world timestamps with dataset-specific discrete time points during the transformation of queries and paths into textual context. To further validate the effectiveness of this approach, we conducted additional experiments by directly querying LLMs using absolute timestamps without our designed discrete time identifiers. The performance results of these experiments are presented in Table~\ref{tab:ab1}. As indicated by the experimental outcomes, the performance of Deepseek-V3 severely deteriorates, becoming nearly unusable. For example, Hits@1 scores on the ICEWS14 and ICEWS0515 dataset drop to merely 5.2\% and 4.6\%, respectively. This performance degradation arises primarily because, without specified discrete timestamps, queries become ambiguous due to numerous similar historical events. For instance, queries such as $'Trump, meet\ with, ?'$ are difficult for LLMs to resolve unambiguously without explicit time constraints. These findings underscore the necessity and effectiveness of employing dataset-specific discrete time points to enforce genuine logical reasoning rather than historical data retrieval in LLM-based temporal knowledge graph reasoning.

\subsubsection{{Computational Efficiency Analysis}}
{A critical concern for LLM-enhanced frameworks is the computational overhead associated with token consumption. We conducted a comparative analysis between IGETR and standard ICL baselines regarding input token volume per query.}

{As reported in previous studies \cite{lee2023temporal}, retrieval-based ICL methods typically necessitate a broad context window that includes the top-100 paths to ensure the ground truth is captured within the noisy retrieval results. Our measurements on the ICEWS14 dataset indicate that this setting results in an average consumption of approximately 3,108 tokens per query.}

{In contrast, IGETR utilizes the temporal GNN as a rigorous structural filter. The attention-guided sampling mechanism allows our model to achieve state-of-the-art performance using only the top-30 candidate paths. Consequently, the average input for IGETR is reduced to approximately 1,758 tokens per query. This represents a 43\% reduction in computational load compared to standard ICL paradigms, demonstrating that the integration of GNNs effectively optimizes the trade-off between reasoning accuracy and computational efficiency.}

\subsection{Case study}

\subsubsection{LLM Hallucination in Temporal Scenarios}
\begin{figure}[t]
	\centering
	\includegraphics[scale=0.48]{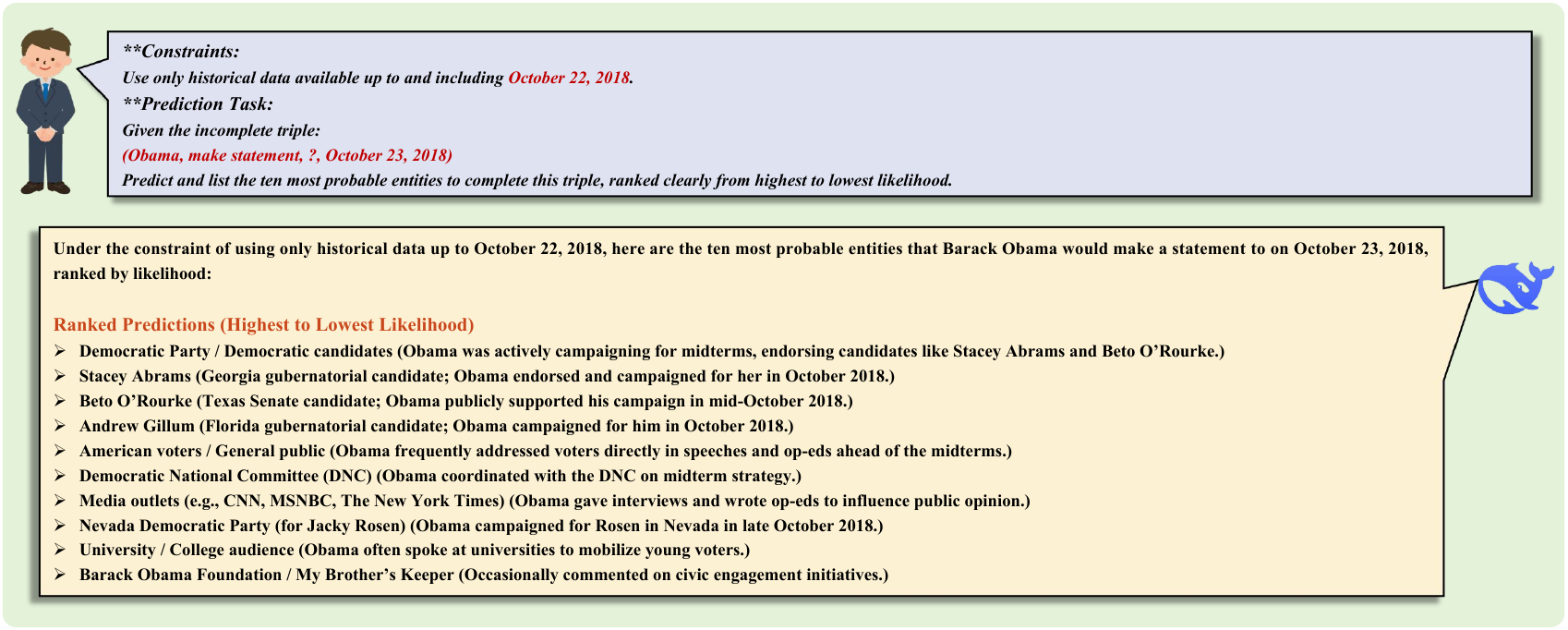}
	\caption{Case study demonstrating LLM hallucination in dynamic temporal reasoning.}
	\label{fig:ab1}
\end{figure}

To highlight the limitations of LLMs under dynamic temporal conditions, we present a case study involving the query: \textit{(Obama, make statement, ?, October 23, 2018)}. {In this experiment, we constrained the model to use only historical information available up to the day before the query. Figure~\ref{fig:ab1} visualizes the outcome of this test. As shown, the LLM failed to identify the correct target entity "Russia" and instead produced predictions centered around prominent but incorrect U.S. political events.} Instead, its predictions predominantly centered around U.S. midterm elections, suggesting figures like Stacey Abrams and Beto O’Rourke. This indicates a tendency to overemphasize prominent patterns from recent political discourse, neglecting less salient but temporally correct facts. Such hallucinations reveal the challenge LLMs face in adapting to evolving knowledge and underscore the need for structured graph guidance in temporal reasoning tasks.

\subsubsection{Improving Paths with LLM}
\begin{figure}[t]
	\centering
	\includegraphics[scale=0.48]{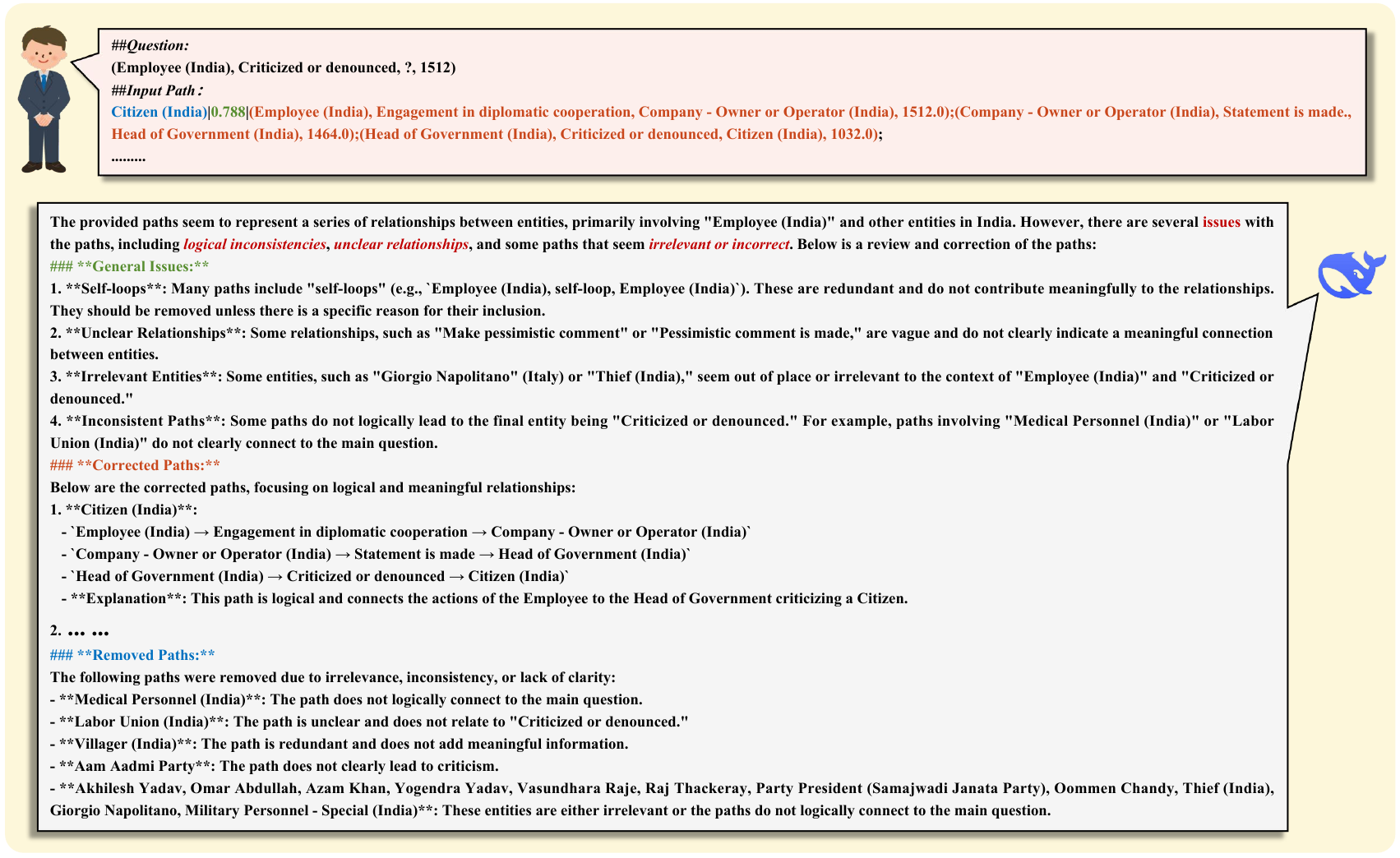}
	\caption{Case study illustrating the effectiveness of the LLM-based path editing module. Given an initial reasoning path extracted by the temporal GNN model (top panel), the LLM systematically identifies and corrects logical inconsistencies, semantic ambiguities, irrelevant entities, and redundant relations (bottom panel).}
	\label{fig:ab2}
\end{figure} 

To intuitively demonstrate the effectiveness of our proposed LLM-based path editing module in improving reasoning paths, we present a detailed case study as depicted in Figure~\ref{fig:ab2}. Given a specific query \textit{(Employee (India), Criticized or denounced, ?, 1512)}, the initial reasoning paths provided by the temporal GNN model include several logically inconsistent, unclear, redundant, and irrelevant relations. Specifically, the original paths exhibit problems such as self-loops, unclear semantics, irrelevant entities, and paths logically disconnected from the target entity.

The LLM-based editing module systematically addresses these issues. First, it identifies and removes redundant paths that contain self-loop relations, as they contribute no meaningful information to the reasoning process. Second, the LLM clarifies ambiguous and vague relations, ensuring that the semantics of the relations explicitly convey meaningful connections. Third, the module effectively filters out irrelevant entities and relations unrelated to the query context, significantly improving the relevance and coherence of the inference chains. Finally, logically inconsistent paths that fail to lead clearly to the queried event are also discarded, enhancing the overall logical consistency.

This case study clearly illustrates that the LLM-based path editing module can efficiently refine reasoning paths at both the semantic and logical levels, resulting in more accurate, interpretable, and contextually coherent inference paths. Thus, the module significantly contributes to the robustness and reliability of our temporal knowledge graph reasoning framework.

\subsection{{Limitations and Failure Analysis}}
{Despite the robust performance of IGETR across aggregate metrics, our qualitative analysis of individual predictions identifies two specific failure modes. First, we observed instances of over-correction where the GNN proposed a factually correct but counter-intuitive reasoning path. Guided by generalized world knowledge, the LLM occasionally misidentified these valid but rare connections as logical errors and altered them into more semantically plausible yet factually incorrect sequences. Second, despite explicit constraints to preserve the original vocabulary, the LLM introduced subtle factual inconsistencies in rare cases by replacing graph entities with external concepts not present in the TKG. However, empirical results indicate that these adverse effects are infrequent. The framework mitigates such risks by instructing the model to respect the confidence scores of the GNN, which prevents the aggressive alteration of paths that are structurally grounded and deemed highly likely to be correct.}

\section{Conclusion}
In this study, we propose IGETR, a novel temporal knowledge graph reasoning framework that integrates the structural modeling strengths of GNNs with the semantic reasoning capabilities of LLMs. By enriching entity and relation embeddings through LLM-enhanced semantic encoding, extracting temporally coherent inference paths via attention-guided GNN, and refining these paths using an LLM-driven editing mechanism, IGETR effectively ensures accurate, logically consistent, and interpretable reasoning. Comprehensive experiments demonstrated that our model consistently outperforms baseline methods on benchmark datasets, highlighting the benefits of combining structured graph information with external semantic knowledge. Ablation and case studies further validated the effectiveness of our path editing module and strategies to prevent unintended historical retrieval by LLMs, underscoring IGETR's potential in reliable, high-stakes temporal reasoning scenarios. {In future, this work opens several promising research directions, such as creating closed-loop systems where the LLM's edits provide feedback to the GNN, or extending this methodology to other complex reasoning domains.}

\section{Declaration of generative AI and AI-assisted technologies in the manuscript preparation process}
During the preparation of this work the authors used Gemini in order to polish the wording and improve the readability of the manuscript. After using this tool/service, the author(s) reviewed and edited the content as needed and take full responsibility for the content of the published article.

 \bibliographystyle{elsarticle-num} 
 \bibliography{ref}

@article{qin2023learning,
  title={Learning from hierarchical structure of knowledge graph for recommendation},
  author={Qin, Yingrong and Gao, Chen and Wei, Shuangqing and Wang, Yue and Jin, Depeng and Yuan, Jian and Zhang, Lin and Li, Dong and Hao, Jianye and Li, Yong},
  journal={ACM Transactions on Information Systems},
  volume={42},
  number={1},
  pages={1--24},
  year={2023},
  publisher={ACM New York, NY, USA}
}

@article{liu2023prompt,
  title={Prompt-WNQA: A prompt-based complex question answering for wireless network over knowledge graph},
  author={Liu, Pei and Qian, Bing and Sun, Qi and Zhao, Longgang},
  journal={Computer Networks},
  volume={236},
  pages={110014},
  year={2023},
  publisher={Elsevier}
}

@ARTICLE{10577554,
  author={Liang, Ke and Meng, Lingyuan and Liu, Meng and Liu, Yue and Tu, Wenxuan and Wang, Siwei and Zhou, Sihang and Liu, Xinwang and Sun, Fuchun and He, Kunlun},
  journal={IEEE Transactions on Pattern Analysis and Machine Intelligence}, 
  title={A Survey of Knowledge Graph Reasoning on Graph Types: Static, Dynamic, and Multi-Modal}, 
  year={2024},
  volume={46},
  number={12},
  pages={9456-9478},
  doi={10.1109/TPAMI.2024.3417451}}

@article{liang2150better,
  title={Better Learning from Graph Structures: Research on Representation Learning for Knowledge Graph Reasoning},
  author={Liang, Ke},
  journal={Proceedings of the VLDB Endowment. ISSN},
  volume={2150},
  pages={8097}
}

@article{ji2021survey,
  title={A survey on knowledge graphs: Representation, acquisition, and applications},
  author={Ji, Shaoxiong and Pan, Shirui and Cambria, Erik and Marttinen, Pekka and Philip, S Yu},
  journal={IEEE transactions on neural networks and learning systems},
  volume={33},
  number={2},
  pages={494--514},
  year={2021},
  publisher={IEEE}
}

@article{cai2024survey,
  title={A survey on temporal knowledge graph: Representation learning and applications},
  author={Cai, Li and Mao, Xin and Zhou, Yuhao and Long, Zhaoguang and Wu, Changxu and Lan, Man},
  journal={arXiv preprint arXiv:2403.04782},
  year={2024}
}

@article{zhang2024can,
  title={Can Large Language Models Improve the Adversarial Robustness of Graph Neural Networks?},
  author={Zhang, Zhongjian and Wang, Xiao and Zhou, Huichi and Yu, Yue and Zhang, Mengmei and Yang, Cheng and Shi, Chuan},
  journal={arXiv preprint arXiv:2408.08685},
  year={2024}
}

@InProceedings{
xerte,
title={Explainable Subgraph Reasoning for Forecasting on Temporal Knowledge Graphs},
author={Zhen Han and Peng Chen and Yunpu Ma and Volker Tresp},
booktitle={International Conference on Learning Representations},
year={2021},
url={https://openreview.net/forum?id=pGIHq1m7PU}
}

@article{fan2024flow,
  title={Flow to Candidate: Temporal Knowledge Graph Reasoning With Candidate-Oriented Relational Graph},
  author={Fan, Shiqi and Fan, Guoxi and Nie, Hongyi and Yao, Quanming and Liu, Yang and Li, Xuelong and Wang, Zhen},
  journal={IEEE Transactions on Neural Networks and Learning Systems},
  year={2024},
  publisher={IEEE}
}

@inproceedings{liang2023learn,
  title={Learn from relational correlations and periodic events for temporal knowledge graph reasoning},
  author={Liang, Ke and Meng, Lingyuan and Liu, Meng and Liu, Yue and Tu, Wenxuan and Wang, Siwei and Zhou, Sihang and Liu, Xinwang},
  booktitle={Proceedings of the 46th international ACM SIGIR conference on research and development in information retrieval},
  pages={1559--1568},
  year={2023}
}

@article{guo2024graphedit,
  title={Graphedit: Large language models for graph structure learning},
  author={Guo, Zirui and Xia, Lianghao and Yu, Yanhua and Wang, Yuling and Yang, Zixuan and Wei, Wei and Pang, Liang and Chua, Tat-Seng and Huang, Chao},
  journal={arXiv preprint arXiv:2402.15183},
  year={2024}
}

@inproceedings{liao2024gentkg,
  title={GenTKG: Generative Forecasting on Temporal Knowledge Graph with Large Language Models},
  author={Liao, Ruotong and Jia, Xu and Li, Yangzhe and Ma, Yunpu and Tresp, Volker},
  booktitle={Findings of the Association for Computational Linguistics: NAACL 2024},
  pages={4303--4317},
  year={2024}
}

@article{luo2023chatrule,
  title={Chatrule: Mining logical rules with large language models for knowledge graph reasoning},
  author={Luo, Linhao and Ju, Jiaxin and Xiong, Bo and Li, Yuan-Fang and Haffari, Gholamreza and Pan, Shirui},
  journal={arXiv preprint arXiv:2309.01538},
  year={2023}
}

@article{ICL2023temporal,
  title={Temporal knowledge graph forecasting without knowledge using in-context learning},
  author={Lee, Dong-Ho and Ahrabian, Kian and Jin, Woojeong and Morstatter, Fred and Pujara, Jay},
  journal={arXiv preprint arXiv:2305.10613},
  year={2023}
}

@article{luo2024chain,
  title={Chain of history: Learning and forecasting with llms for temporal knowledge graph completion},
  author={Luo, Ruilin and Gu, Tianle and Li, Haoling and Li, Junzhe and Lin, Zicheng and Li, Jiayi and Yang, Yujiu},
  journal={arXiv preprint arXiv:2401.06072},
  year={2024}
}

@article{pan2024unifying,
  title={Unifying large language models and knowledge graphs: A roadmap},
  author={Pan, Shirui and Luo, Linhao and Wang, Yufei and Chen, Chen and Wang, Jiapu and Wu, Xindong},
  journal={IEEE Transactions on Knowledge and Data Engineering},
  volume={36},
  number={7},
  pages={3580--3599},
  year={2024},
  publisher={IEEE}
}

@article{dwivedi2023explainable,
  title={Explainable AI (XAI): Core ideas, techniques, and solutions},
  author={Dwivedi, Rudresh and Dave, Devam and Naik, Het and Singhal, Smiti and Omer, Rana and Patel, Pankesh and Qian, Bin and Wen, Zhenyu and Shah, Tejal and Morgan, Graham and others},
  journal={ACM Computing Surveys},
  volume={55},
  number={9},
  pages={1--33},
  year={2023},
  publisher={ACM New York, NY}
}

@article{han2024parameter,
  title={Parameter-efficient fine-tuning for large models: A comprehensive survey},
  author={Han, Zeyu and Gao, Chao and Liu, Jinyang and Zhang, Jeff and Zhang, Sai Qian},
  journal={arXiv preprint arXiv:2403.14608},
  year={2024}
}

@article{gao2023retrieval,
  title={Retrieval-augmented generation for large language models: A survey},
  author={Gao, Yunfan and Xiong, Yun and Gao, Xinyu and Jia, Kangxiang and Pan, Jinliu and Bi, Yuxi and Dai, Yi and Sun, Jiawei and Wang, Haofen and Wang, Haofen},
  journal={arXiv preprint arXiv:2312.10997},
  volume={2},
  year={2023}
}

@inproceedings{li2021temporal,
  title={Temporal knowledge graph reasoning based on evolutional representation learning},
  author={Li, Zixuan and Jin, Xiaolong and Li, Wei and Guan, Saiping and Guo, Jiafeng and Shen, Huawei and Wang, Yuanzhuo and Cheng, Xueqi},
  booktitle={Proceedings of the 44th international ACM SIGIR conference on research and development in information retrieval},
  pages={408--417},
  year={2021}
}

@article{xu2020inductive,
  title={Inductive representation learning on temporal graphs},
  author={Xu, Da and Ruan, Chuanwei and Korpeoglu, Evren and Kumar, Sushant and Achan, Kannan},
  journal={arXiv preprint arXiv:2002.07962},
  year={2020}
}

@inproceedings{xu2022ruleformer,
  title={Ruleformer: Context-aware rule mining over knowledge graph},
  author={Xu, Zezhong and Ye, Peng and Chen, Hui and Zhao, Meng and Chen, Huajun and Zhang, Wen},
  booktitle={Proceedings of the 29th International Conference on Computational Linguistics},
  pages={2551--2560},
  year={2022}
}

@inproceedings{jin2020recurrent,
  title={Recurrent Event Network: Autoregressive Structure Inferenceover Temporal Knowledge Graphs},
  author={Jin, Woojeong and Qu, Meng and Jin, Xisen and Ren, Xiang},
  booktitle={Proceedings of the 2020 Conference on Empirical Methods in Natural Language Processing (EMNLP)},
  pages={6669--6683},
  year={2020}
}

@inproceedings{han2021learning,
  title={Learning neural ordinary equations for forecasting future links on temporal knowledge graphs},
  author={Han, Zhen and Ding, Zifeng and Ma, Yunpu and Gu, Yujia and Tresp, Volker},
  booktitle={Proceedings of the 2021 conference on empirical methods in natural language processing},
  pages={8352--8364},
  year={2021}
}

@inproceedings{li2022tirgn,
  title={TiRGN: Time-Guided Recurrent Graph Network with Local-Global Historical Patterns for Temporal Knowledge Graph Reasoning.},
  author={Li, Yujia and Sun, Shiliang and Zhao, Jing},
  booktitle={IJCAI},
  pages={2152--2158},
  year={2022}
}

@inproceedings{zhang2023learning,
  title={Learning long-and short-term representations for temporal knowledge graph reasoning},
  author={Zhang, Mengqi and Xia, Yuwei and Liu, Qiang and Wu, Shu and Wang, Liang},
  booktitle={Proceedings of the ACM web conference 2023},
  pages={2412--2422},
  year={2023}
}

@inproceedings{chu2024timebench,
  title={TimeBench: A Comprehensive Evaluation of Temporal Reasoning Abilities in Large Language Models},
  author={Chu, Zheng and Chen, Jingchang and Chen, Qianglong and Yu, Weijiang and Wang, Haotian and Liu, Ming and Qin, Bing},
  booktitle={Proceedings of the 62nd Annual Meeting of the Association for Computational Linguistics (Volume 1: Long Papers)},
  pages={1204--1228},
  year={2024}
}

@inproceedings{luoreasoning,
  title={Reasoning on Graphs: Faithful and Interpretable Large Language Model Reasoning},
  author={LUO, LINHAO and Li, Yuan-Fang and Haf, Reza and Pan, Shirui},
  booktitle={The Twelfth International Conference on Learning Representations}
}

@inproceedings{xu2023pre,
  title={Pre-trained Language Model with Prompts for Temporal Knowledge Graph Completion},
  author={Xu, Wenjie and Liu, Ben and Peng, Miao and Jia, Xu and Peng, Min},
  booktitle={Findings of the Association for Computational Linguistics: ACL 2023},
  pages={7790--7803},
  year={2023}
}

@inproceedings{lee2023temporal,
  title={Temporal Knowledge Graph Forecasting Without Knowledge Using In-Context Learning},
  author={Lee, Dong-Ho and Ahrabian, Kian and Jin, Woojeong and Morstatter, Fred and Pujara, Jay},
  booktitle={Proceedings of the 2023 Conference on Empirical Methods in Natural Language Processing},
  pages={544--557},
  year={2023}
}

@article{guo2023gpt4graph,
  title={Gpt4graph: Can large language models understand graph structured data? an empirical evaluation and benchmarking},
  author={Guo, Jiayan and Du, Lun and Liu, Hengyu and Zhou, Mengyu and He, Xinyi and Han, Shi},
  journal={arXiv preprint arXiv:2305.15066},
  year={2023}
}

@article{chai2023graphllm,
  title={Graphllm: Boosting graph reasoning ability of large language model},
  author={Chai, Ziwei and Zhang, Tianjie and Wu, Liang and Han, Kaiqiao and Hu, Xiaohai and Huang, Xuanwen and Yang, Yang},
  journal={arXiv preprint arXiv:2310.05845},
  year={2023}
}

@data{DVN/28075_2015,
author = {Boschee, Elizabeth and Lautenschlager, Jennifer and O'Brien, Sean and Shellman, Steve and Starz, James and Ward, Michael},
publisher = {Harvard Dataverse},
title = {{ICEWS Coded Event Data}},
UNF = {UNF:6:NOSHB7wyt0SQ8sMg7+w38w==},
year = {2015},
version = {V30},
doi = {10.7910/DVN/28075},
url = {https://doi.org/10.7910/DVN/28075}
}

@article{guo2025semantic,
  title={Semantic information-based attention mapping network for few-shot knowledge graph completion},
  author={Guo, Fan and Chang, Xiangmao and Guo, Yunqi and Xing, Guoliang and Zhao, Yunlong},
  journal={Neural Networks},
  pages={107366},
  year={2025},
  publisher={Elsevier}
}

@article{zhang2023few,
  title={Few-shot link prediction for temporal knowledge graphs based on time-aware translation and attention mechanism},
  author={Zhang, Han and Bai, Luyi},
  journal={Neural Networks},
  volume={161},
  pages={371--381},
  year={2023},
  publisher={Elsevier}
}

@article{li2025temporal,
  title={Temporal multi-modal knowledge graph generation for link prediction},
  author={Li, Yuandi and Ji, Hui and Yu, Fei and Cheng, Lechao and Che, Nan},
  journal={Neural Networks},
  volume={185},
  pages={107108},
  year={2025},
  publisher={Elsevier}
}

@article{feng2025retrieval,
  title={Retrieval In Decoder benefits generative models for explainable complex question answering},
  author={Feng, Jianzhou and Wang, Qin and Qiu, Huaxiao and Liu, Lirong},
  journal={Neural Networks},
  volume={181},
  pages={106833},
  year={2025},
  publisher={Elsevier}
}

@article{maltoni2024arithmetic,
  title={Arithmetic with language models: From memorization to computation},
  author={Maltoni, Davide and Ferrara, Matteo},
  journal={Neural Networks},
  volume={179},
  pages={106550},
  year={2024},
  publisher={Elsevier}
}

@article{yang2025language,
  title={Language-based reasoning graph neural network for commonsense question answering},
  author={Yang, Meng and Wang, Yihao and Gu, Yu},
  journal={Neural Networks},
  volume={181},
  pages={106816},
  year={2025},
  publisher={Elsevier}
}

@article{huang2024knowledge,
  title={Knowledge graph confidence-aware embedding for recommendation},
  author={Huang, Chen and Yu, Fei and Wan, Zhiguo and Li, Fengying and Ji, Hui and Li, Yuandi},
  journal={Neural Networks},
  pages={106601},
  year={2024},
  publisher={Elsevier}
}

@article{dai2022mrgat,
  title={MRGAT: multi-relational graph attention network for knowledge graph completion},
  author={Dai, Guoquan and Wang, Xizhao and Zou, Xiaoying and Liu, Chao and Cen, Si},
  journal={Neural Networks},
  volume={154},
  pages={234--245},
  year={2022},
  publisher={Elsevier}
}

@article{ma2024harnessing,
  title={Harnessing collective structure knowledge in data augmentation for graph neural networks},
  author={Ma, Rongrong and Pang, Guansong and Chen, Ling},
  journal={Neural Networks},
  volume={180},
  pages={106651},
  year={2024},
  publisher={Elsevier}
}

@InProceedings{ttranse,
  title={Deriving validity time in knowledge graph},
  author={Leblay, Julien and Chekol, Melisachew Wudage},
  booktitle={Companion Proceedings of the The Web Conference 2018},
  pages={1771--1776},
  year={2018}
}

@InProceedings{
TNTCOMP,
title={Tensor Decompositions for Temporal Knowledge Base Completion},
author={Timothée Lacroix and Guillaume Obozinski and Nicolas Usunier},
booktitle={International Conference on Learning Representations},
year={2020},
url={https://openreview.net/forum?id=rke2P1BFwS}
}

@article{ChronoR, 
title={ChronoR: Rotation Based Temporal Knowledge Graph Embedding}, volume={35}, url={https://ojs.aaai.org/index.php/AAAI/article/view/16802}, abstractNote={Despite the importance and abundance of temporal knowledge graphs, most of the current research has been focused on reasoning on static graphs. In this paper, we study the challenging problem of inference over temporal knowledge graphs. In particular, the task of temporal link prediction. In general, this is a difficult task due to data non-stationarity, data heterogeneity, and its complex temporal dependencies. We propose Chronological Rotation embedding (ChronoR), a novel model for learning representations for entities, relations, and time. Learning dense representations is frequently used as an efficient and versatile method to perform reasoning on knowledge graphs. The proposed model learns a k-dimensional rotation transformation parametrized by relation and time, such that after each fact’s head entity is transformed using the rotation, it falls near its corresponding tail entity. By using high dimensional rotation as its transformation operator, ChronoR captures rich interaction between the temporal and multi-relational characteristics of a Temporal Knowledge Graph. Experimentally, we show that ChronoR is able to outperform many of the state-of-the-art methods on the benchmark datasets for temporal knowledge graph link prediction.}, number={7}, journal={Proceedings of the AAAI Conference on Artificial Intelligence}, author={Sadeghian, Ali and Armandpour, Mohammadreza and Colas, Anthony and Wang, Daisy Zhe}, year={2021}, month={May}, pages={6471-6479} }

@InProceedings{xu-etal-2021-telm,
    title = "Temporal Knowledge Graph Completion using a Linear Temporal Regularizer and Multivector Embeddings",
    author = "Xu, Chengjin  and
      Chen, Yung-Yu  and
      Nayyeri, Mojtaba  and
      Lehmann, Jens",
    booktitle = "Proceedings of the 2021 Conference of the North American Chapter of the Association for Computational Linguistics: Human Language Technologies",
    month = jun,
    year = "2021",
    address = "Online",
    publisher = "Association for Computational Linguistics",
    url = "https://aclanthology.org/2021.naacl-main.202",
    doi = "10.18653/v1/2021.naacl-main.202",
    pages = "2569--2578",
}

@InProceedings{han-etal-2021-tango,
    title = "Learning Neural Ordinary Equations for Forecasting Future Links on Temporal Knowledge Graphs",
    author = "Han, Zhen  and
      Ding, Zifeng  and
      Ma, Yunpu  and
      Gu, Yujia  and
      Tresp, Volker",
    booktitle = "Proceedings of the 2021 Conference on Empirical Methods in Natural Language Processing",
    month = nov,
    year = "2021",
    address = "Online and Punta Cana, Dominican Republic",
    publisher = "Association for Computational Linguistics",
    url = "https://aclanthology.org/2021.emnlp-main.658",
    doi = "10.18653/v1/2021.emnlp-main.658",
    pages = "8352--8364",
}

@inproceedings{jiang2023structgpt,
  title={StructGPT: A General Framework for Large Language Model to Reason over Structured Data},
  author={Jiang, Jinhao and Zhou, Kun and Dong, Zican and Ye, Keming and Zhao, Wayne Xin and Wen, Ji-Rong},
  booktitle={Proceedings of the 2023 Conference on Empirical Methods in Natural Language Processing},
  pages={9237--9251},
  year={2023}
}

@article{zhang2023instruction,
  title={Instruction tuning for large language models: A survey},
  author={Zhang, Shengyu and Dong, Linfeng and Li, Xiaoya and Zhang, Sen and Sun, Xiaofei and Wang, Shuhe and Li, Jiwei and Hu, Runyi and Zhang, Tianwei and Wu, Fei and others},
  journal={arXiv preprint arXiv:2308.10792},
  year={2023}
}

@inproceedings{wei2023kicgpt,
  title={KICGPT: Large Language Model with Knowledge in Context for Knowledge Graph Completion},
  author={Wei, Yanbin and Huang, Qiushi and Zhang, Yu and Kwok, James},
  booktitle={Findings of the Association for Computational Linguistics: EMNLP 2023},
  pages={8667--8683},
  year={2023}
}

@inproceedings{wang2024kc,
  title={KC-GenRe: A Knowledge-constrained Generative Re-ranking Method Based on Large Language Models for Knowledge Graph Completion},
  author={Wang, Yilin and Hu, Minghao and Huang, Zhen and Li, Dongsheng and Yang, Dong and Lu, Xicheng},
  booktitle={Proceedings of the 2024 Joint International Conference on Computational Linguistics, Language Resources and Evaluation (LREC-COLING 2024)},
  pages={9668--9680},
  year={2024}
}

@inproceedings{xu2024multi,
  title={Multi-perspective Improvement of Knowledge Graph Completion with Large Language Models},
  author={Xu, Derong and Zhang, Ziheng and Lin, Zhenxi and Wu, Xian and Zhu, Zhihong and Xu, Tong and Zhao, Xiangyu and Zheng, Yefeng and Chen, Enhong},
  booktitle={Proceedings of the 2024 Joint International Conference on Computational Linguistics, Language Resources and Evaluation (LREC-COLING 2024)},
  pages={11956--11968},
  year={2024}
}

@article{shu2024knowledge,
  title={Knowledge Graph Large Language Model (KG-LLM) for Link Prediction:(ACML)},
  author={Shu, Dong and Chen, Tianle and Jin, Mingyu and Zhang, Chong and Du, Mengnan and Zhang, Yongfeng},
  journal={Proceedings of Machine Learning Research},
  volume={260},
  number={1},
  pages={143},
  year={2024},
  publisher={ML Research Press}
}

@inproceedings{wang2024llm,
  title={LLM as Prompter: Low-resource Inductive Reasoning on Arbitrary Knowledge Graphs},
  author={Wang, Kai and Xu, Yuwei and Wu, Zhiyong and Luo, Siqiang},
  booktitle={Findings of the Association for Computational Linguistics ACL 2024},
  pages={3742--3759},
  year={2024}
}

@inproceedings{liu2022tlogic,
  title={Tlogic: Temporal logical rules for explainable link forecasting on temporal knowledge graphs},
  author={Liu, Yushan and Ma, Yunpu and Hildebrandt, Marcel and Joblin, Mitchell and Tresp, Volker},
  booktitle={Proceedings of the AAAI conference on artificial intelligence},
  volume={36},
  number={4},
  pages={4120--4127},
  year={2022}
}

@article{A,
  title={A resource-aware multi-graph neural network for urban traffic flow prediction in multi-access edge computing systems},
  author={Ali, Ahmad and Ullah, Inam and Shabaz, Mohammad and Sharafian, Amin and Khan, Muhammad Attique and Bai, Xiaoshan and Qiu, Li},
  journal={IEEE Transactions on Consumer Electronics},
  year={2024},
  publisher={IEEE}
}

@article{B,
  title={Exploiting dynamic spatio-temporal graph convolutional neural networks for citywide traffic flows prediction},
  author={Ali, Ahmad and Zhu, Yanmin and Zakarya, Muhammad},
  journal={Neural networks},
  volume={145},
  pages={233--247},
  year={2022},
  publisher={Elsevier}
}

@inproceedings{C,
  title={VMR: virtual machine replacement algorithm for QoS and energy-awareness in cloud data centers},
  author={Ali, Riaz and Shen, Yao and Huang, Xiangwei and Zhang, Jingyu and Ali, Ahmad},
  booktitle={2017 IEEE International Conference on Computational Science and Engineering (CSE) and IEEE International Conference on Embedded and Ubiquitous Computing (EUC)},
  volume={2},
  pages={230--233},
  year={2017},
  organization={IEEE}
}

@article{D,
  title={An attention-driven spatio-temporal deep hybrid neural networks for traffic flow prediction in transportation systems},
  author={Ali, Ahmad and Ullah, Inam and Ahmad, Shabir and Wu, Zongze and Li, Jianqiang and Bai, Xiaoshan},
  journal={IEEE Transactions on Intelligent Transportation Systems},
  year={2025},
  publisher={IEEE}
}

@article{E,
  title={PH-GCN: Boosting Human Action Recognition through Multi-Level Granularity with Pair-wise Hyper GCN},
  author={Alsarhan, Tamam and Ali, Syed Sadaf and Ganapathi, Iyyakutti Iyappan and Ali, Ahmad and Werghi, Naoufel},
  journal={IEEE Access},
  year={2024},
  publisher={IEEE}
}

@article{F,
  title={Energy-Efficient Resource Allocation for Urban Traffic Flow Prediction in Edge-Cloud Computing},
  author={Ali, Ahmad and Ullah, Inam and Singh, Sushil Kumar and Sharafian, Amin and Jiang, Weiwei and I. Sherazi, Hammad and Bai, Xiaoshan},
  journal={International Journal of Intelligent Systems},
  volume={2025},
  number={1},
  pages={1863025},
  year={2025},
  publisher={Wiley Online Library}
}

@article{G,
  title={Attention-Driven Graph Convolutional Networks for Deadline-Constrained Virtual Machine Task Allocation in Edge Computing},
  author={Ali, Ahmad and Ullah, Inam and Singh, Sushil Kumar and Jiang, Weiwei and Alturise, Fahad and Bai, Xiaoshan},
  journal={IEEE Transactions on Consumer Electronics},
  year={2025},
  publisher={IEEE}
}

@article{RefA,
  title={Improvement of energy-efficient resources for cognitive internet of things using learning automata},
  author={Rahmani, Parisa and Arefi, Mohamad},
  journal={Peer-to-Peer Networking and Applications},
  volume={17},
  number={1},
  pages={297--320},
  year={2024},
  publisher={Springer}
}

@article{RefB,
  title={IoT-RNNEI: An Internet of Things Attack Detection Model Leveraging Random Neural Network and Evolutionary Intelligence},
  author={Rahmani, Parisa and Arefi, Mohamad and Shojae, Seyyed Mohammad Saber Seyyed and Mirzaee, Ashraf},
  journal={IET Communications},
  volume={19},
  number={1},
  pages={e70055},
  year={2025},
  publisher={Wiley Online Library}
}

@inproceedings{xia2024chain,
  title={Chain-of-History Reasoning for Temporal Knowledge Graph Forecasting},
  author={Xia, Yuwei and Wang, Ding and Liu, Qiang and Wang, Liang and Wu, Shu and Zhang, Xiao-Yu},
  booktitle={Findings of the Association for Computational Linguistics ACL 2024},
  pages={16144--16159},
  year={2024}
}






\end{document}